\documentclass[acmtog]{acmart}

\usepackage{booktabs} %

\usepackage{color,xcolor}
\usepackage{graphicx}

\usepackage{microtype}
\usepackage[font=small]{caption}
\usepackage{arydshln}
\usepackage{tabularx}
\usepackage{array}
\usepackage{booktabs}
\usepackage{colortbl}
\usepackage{float,wrapfig}
\usepackage{hhline}
\usepackage{multirow}
\usepackage{subcaption} %

\usepackage{duckuments}
\usepackage{amsmath}

\usepackage{bm}
\usepackage{nicefrac}
\usepackage{microtype}
\usepackage{dsfont}
\usepackage{changepage}
\usepackage{extramarks}
\usepackage{fancyhdr}
\usepackage{setspace}
\usepackage{soul}
\usepackage{xspace}
\usepackage{hhline}
\usepackage{algorithmicx}
\usepackage{algpseudocode}
\usepackage{pifont}
\usepackage{booktabs}
\usepackage{multirow}

\usepackage{makecell}

\usepackage{enumitem}
\usepackage[title]{appendix}

\usepackage[ruled]{algorithm2e} %

\SetAlFnt{\small}
\SetAlCapFnt{\small}
\SetAlCapNameFnt{\small}
\SetAlCapHSkip{0pt}
\IncMargin{-\parindent}

\setcopyright{cc}

\renewcommand\footnotetextcopyrightpermission[1]{} %
\acmConference[SA Conference Papers '23]{SIGGRAPH Asia 2023 Conference Papers}{December 12--15, 2023}{Sydney, NSW, Australia}

\setcitestyle{square}

\usepackage{amsmath,amsfonts,bm,multirow,}
\usepackage{duckuments}
\usepackage{arydshln}
\usepackage{xcolor}

\definecolor{MyDarkBlue}{rgb}{0,0.08,1}
\definecolor{MyDarkGreen}{rgb}{0.02,0.6,0.02}
\definecolor{MyDarkRed}{rgb}{0.8,0.02,0.02}
\definecolor{MyDarkOrange}{rgb}{0.40,0.2,0.02}
\definecolor{MyPurple}{RGB}{111,0,255}
\definecolor{MyRed}{rgb}{1.0,0.0,0.0}
\definecolor{MyGold}{rgb}{0.75,0.6,0.12}
\definecolor{MyDarkgray}{rgb}{0.66, 0.66, 0.66}
\definecolor{MyDarkCyan}{rgb}{0.05, 0.55, 0.55}
\definecolor{MyBlack}{rgb}{0., 0., 0.}
\definecolor{MyMagenta}{rgb}{1., 0., 1.}
\definecolor{BerkeleyYellow}{RGB}{255,204,41}
\definecolor{BerkeleyLightBlue}{RGB}{94,146,221}
\definecolor{BkDarkBlue}{rgb}{.05,.07,.353}

\definecolor{MyDarkGray2}{rgb}{0.6, 0.6, 0.6}

\newcommand\blfootnote[1]{%
  \begingroup
  \renewcommand\thefootnote{}\footnote{#1}%
  \addtocounter{footnote}{-1}%
  \endgroup
}
\newcommand{\sysubmit}[1]{#1}

\newcommand{\camready}[1]{#1}

\newcommand{\dinorm}[1]{#1}

\newcommand{\suppmat}[1]{#1}

\newcommand{\g}{\theta}
\newcommand{\mI}{N}
\newcommand{\mi}{n}
\newcommand{\gcolle}{\g\sim \text{unif}\{\g_1, \g_2, \dots, \g_\mI\}}
\newcommand{\gset}{\g\in \{\g_1, \dots, \g_\mI\}}
\newcommand{\gsett}{n \in \{1, \dots, \mI\}}
\newcommand{\im}{x}
\newcommand{\gi}{\g}
\newcommand{\qry}{q}
\newcommand{\imdist}{p(x|\theta)}
\newcommand{\mdlprob}{p(\g | \qry)}

\newcommand{\imfeat}{z}
\newcommand{\imclip}{h_\im}
\newcommand{\txtclip}{h}
\newcommand{\clipimex}{\phi_\text{im}}
\newcommand{\cliptxtex}{\phi_\text{txt}}
\newcommand{\nmimclip}{\Tilde{\imclip}}
\newcommand{\nmtxtclip}{\Tilde{\txtclip_\qry}}

\newcommand{\gauden}{Gaussian Density}
\newcommand{\mc}{Monte-Carlo}
\newcommand{\firstmom}{1\textsuperscript{st} Moment}

\newcommand{\reffig}[1]{Figure~\ref{fig:#1}}
\newcommand{\refsec}[1]{Section~\ref{sec:#1}}
\newcommand{\refapp}[1]{Appendix~\ref{sec:#1}}
\newcommand{\reftbl}[1]{Table~\ref{tbl:#1}}

\newcommand{\refeq}[1]{Equation~\ref{eq:#1}}

\newcommand{\lblfig}[1]{\label{fig:#1}}
\newcommand{\lblsec}[1]{\label{sec:#1}}
\newcommand{\lbleq}[1]{\label{eq:#1}}
\newcommand{\lbltbl}[1]{\label{tbl:#1}}

\newcommand{\ignorethis}[1]{}

\newcommand{\myparagraph}[1]{\vspace{-3pt}\paragraph{#1}}

\def\1{\bm{1}}

\newcommand{\ignore}[1]{}

\renewcommand*{\thefootnote}{\arabic{footnote}}

\makeatletter
\DeclareRobustCommand\onedot{\futurelet\@let@token\@onedot}
\def\@onedot{\ifx\@let@token.\else.\null\fi\xspace}

\makeatother

\begin{document}
\title{Content-Based Search for Deep Generative Models}
\author{Daohan Lu*}
\orcid{0000-0002-8733-6177}
\affiliation{%
 \institution{Carnegie Mellon University}
 \streetaddress{5000 Forbes Ave}
 \city{Pittsburgh}
 \state{PA}
 \postcode{15213}
 \country{USA}
}
\email{dl3957@nyu.edu}

\author{Sheng-Yu Wang*}
\orcid{0000-0003-4000-2046}
\affiliation{%
 \institution{Carnegie Mellon University}
 \streetaddress{5000 Forbes Ave}
 \city{Pittsburgh}
 \state{PA}
 \postcode{15213}
 \country{USA}
}
\email{shengyu2@andrew.cmu.edu}

\author{Nupur Kumari*}
\orcid{0000-0003-1799-1069}
\affiliation{%
 \institution{Carnegie Mellon University}
 \streetaddress{5000 Forbes Ave}
 \city{Pittsburgh}
 \state{PA}
 \postcode{15213}
 \country{USA}
}
\email{nkumari@andrew.cmu.edu}

\author{Rohan Agarwal*}
\orcid{0009-0000-3525-5765}
\affiliation{%
 \institution{Carnegie Mellon University}
 \streetaddress{5000 Forbes Ave}
 \city{Pittsburgh}
 \state{PA}
 \postcode{15213}
 \country{USA}
}
\email{agarwal.rohan96@gmail.com}

\author{Mia Tang}
\orcid{0009-0009-3553-3732}
\affiliation{%
 \institution{Carnegie Mellon University}
 \streetaddress{5000 Forbes Ave}
 \city{Pittsburgh}
 \state{PA}
 \postcode{15213}
 \country{USA}
}
\email{miatang13@gmail.com}

\author{David Bau}
\orcid{0000-0003-1744-6765}
\affiliation{%
\institution{Northeastern University}
\streetaddress{440 Huntington Ave}
\city{Boston}
\state{MA}
\postcode{02115}
\country{USA}}
\email{davidbau@northeastern.edu}

\author{Jun-Yan Zhu}
\orcid{0000-0001-8504-3410}
\affiliation{%
 \institution{Carnegie Mellon University}
 \streetaddress{5000 Forbes Ave}
 \city{Pittsburgh}
 \state{PA}
 \postcode{15213}
 \country{USA}
}
\email{junyanz@andrew.cmu.edu}

\renewcommand{\shortauthors}{Lu, Wang, Kumari, Agarwal, Tang, Bau, and Zhu.}

\begin{abstract}
The growing proliferation of customized and pretrained generative models has made it infeasible for a user to be fully cognizant of every model in existence. To address this need, we introduce the task of \emph{content-based model search}: given a query and a large set of generative models, finding the models that best match the query. As each generative model produces a distribution of images, we formulate the search task as an optimization problem to select the model with the highest probability of generating similar content as the query. 
We introduce a formulation to approximate this probability given the query from different modalities, e.g., image, sketch, and text. Furthermore, we propose a contrastive learning framework for model retrieval, which learns to adapt features for various query modalities. We demonstrate that our method outperforms several baselines on \emph{Generative Model Zoo}, a new benchmark we create for the model retrieval task. 

\end{abstract}

\begin{teaserfigure}
    \centering
    \includegraphics[width=\linewidth]{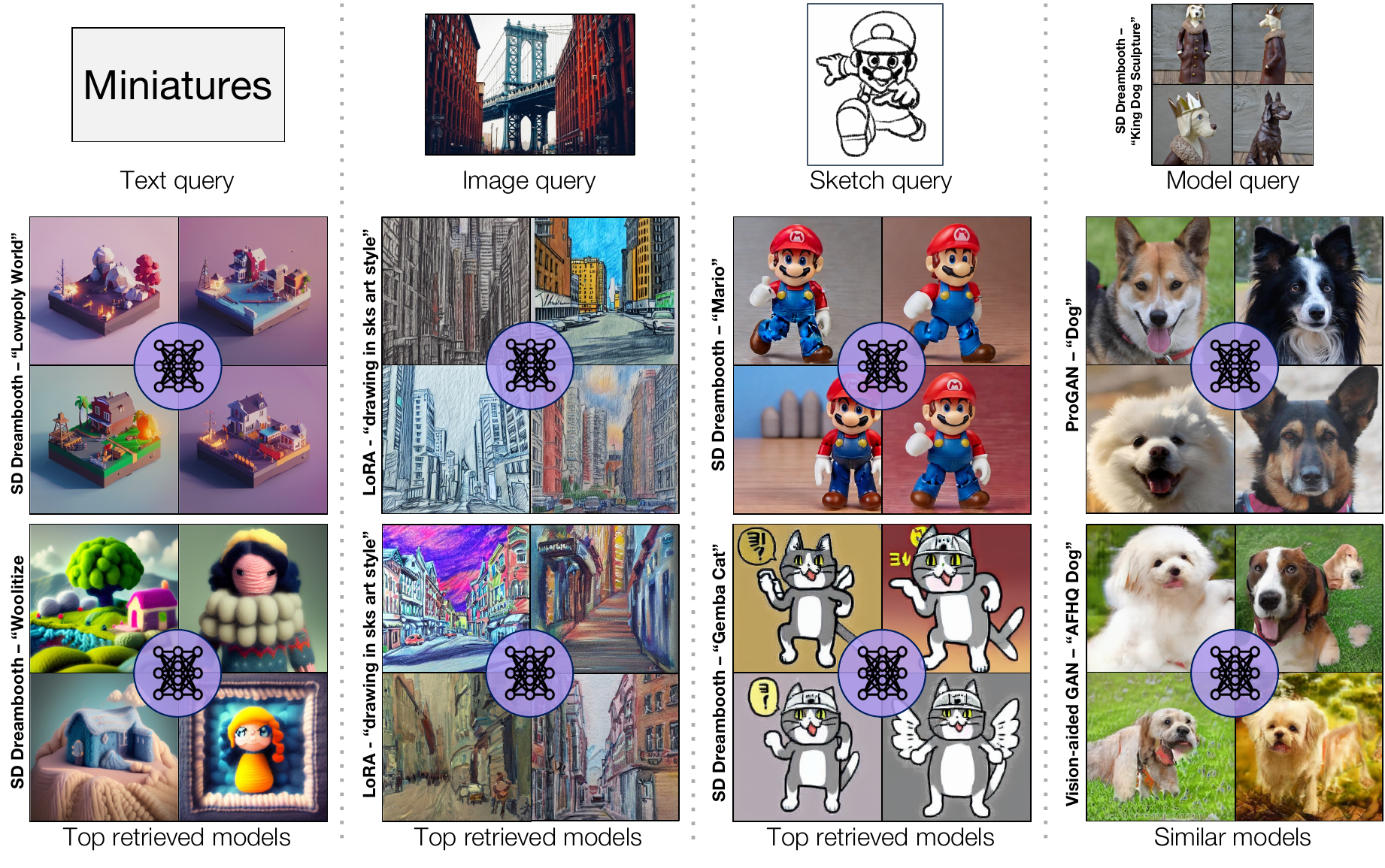}
    \vspace{-18pt}
    \caption{We develop a search algorithm for customized and pre-trained deep generative models. We collect a diverse set of models, such as animals, landscapes, human faces, and art pieces. From left to right, our search algorithm enables queries in four modalities: text, images, sketches, and existing models. The 1\textsuperscript{st} row consists of the queries, and the 2\textsuperscript{nd} and 3\textsuperscript{rd} rows show the first- and second-ranked models by our method, respectively. Our method succeeds in finding relevant models with similar semantic concepts across all four modalities. Different colors denote different model types. \camready{Photograph by \href{https://pixabay.com/users/12019-12019/}{@12019} on \href{https://pixabay.com/photos/george-washington-bridge-2098351/}{Pixabay}.}}
    \label{fig:teaser}
\end{teaserfigure}

\maketitle
\blfootnote{\small{* indicates equal contribution}}
\section{Introduction}
\lblsec{intro}
We introduce the task of content-based model search, which aims to find the most relevant deep image generative models that satisfy a user's input query.
For example, as shown in \reffig{teaser}, we enable a user to retrieve a model capable of synthesizing images that match a  text query (e.g., \texttt{Miniatures}),  an image query (e.g., a landscape photo), sketch query (e.g., Mario sketch), or models similar to a given model. Users can leverage retrieved models in various ways: they might further fine-tune it with a unique concept, use the model to generate an image with a more concrete prompt, or combine multiple retrieved models into one. %

Content-based model search is becoming a vital task as the number and diversity of intriguing generative models continue to explode. These models have formed a new form of media content that requires efficient indexing and searching. %
While one might have expected large-scale text-to-image models~\cite{rombach2021highresolution,ramesh2022hierarchical,saharia2022photorealistic} to have led to a shrinking universe of useful models, instead, this powerful class of content generators has touched off an accelerating proliferation of new and customized models.  That is because powerful generative models are capable of being adapted to capture a wide range of personalized subjects~\cite{kawar2022imagic,ruiz2022dreambooth,kumari2022multi,gal2023designing}. Everyday users are routinely creating new generative models, and the community has collectively shared \emph{tens of thousands} of custom models on community-driven platforms, such as Civitai~\cite{civitailink} and HuggingFace~\cite{huggingfacehub} in the past year alone. Many popular models arise from the model creators' unique creative processes, involving careful selections of subjects, algorithms, hyperparameters, and often proprietary artistic training data. Furthermore, some model creators receive compensation when other users download or use their models.

Aside from personalized models, a variety of deep generative models are being created as backbones for computer graphics applications~\cite{Tewari2020NeuralSTAR,bermano2022state}, and as works of art that explore a wide range of themes~\cite{elgammal2019ai,hertzmann2020computers}. Each model captures a small universe of curated subjects, which can range from the realistic rendering of faces and landscapes~\cite{Karras2021alias} to photos of historic pottery~\cite{au2019vesselgan} to cartoon caricatures~\cite{jang2021stylecari} to single-artist stylistic elements~\cite{schultz2020freagan}. Various methods also enable creative modifications of existing models via human-in-the-loop interfaces~\cite{bau2020rewriting,gal2021stylegan,wang2022rewriting}. %

Given the accelerating pace at which generative models are being created and uploaded to the internet, the ability to conduct content-based model search will be instrumental in supporting the effective use of large model collections. Existing model-sharing platforms~\cite{huggingfacehub,civitailink}  primarily rely on matching human-created model names for model search. However, it is difficult to choose sufficiently detailed names to describe the unique, complex, and highly specific images produced by a generative model. Our work aims to define a new approach, searching for models directly based on their content rather than based on manual definitions alone.

Content-based model search is a challenging task: even the simplified question of whether a specific image can be produced by a single model can be computationally difficult. Unfortunately, many generative models do not offer an efficient or exact way to estimate density, nor do they natively support assessing cross-modal similarity (e.g., text and image). %

To address the above challenges,  we first curate a benchmark retrieval dataset, the Generative Model Zoo, consisting of (1) 259 publicly available generative models that vary in content as well as model architectures, including GANs (e.g., StyleGAN-family models~\cite{karras2020analyzing}), diffusion models (e.g., DreamBooth~\cite{ruiz2022dreambooth}), and auto-regressive models (e.g., VQGAN~\cite{esser2021taming}) and (2) 1000 customized text-to-image diffusion models, each one fine-tuned on a single instantiation of an object class or a single artistic image. As part of our benchmark, we define ground truth image, text, and sketch queries for evaluating model retrieval.

We present a general probabilistic formulation of the model search problem and propose a learning-based method given this formulation. We summarize our contribution as follows:
\begin{itemize}[topsep=0pt]
  \item We introduce a new task of content-based search over deep generative models.  Given a piece of text, an image, a sketch, a generative model, or a combination of them as a query, we aim to return the most relevant generative models that can synthesize similar content.

  \item We formulate the task of content-based model retrieval as estimating the probability of generating an image that matches the query content. We propose a new contrastive learning method to estimate the same given the model's image distribution statistics and query. Our learning-based method outperforms several baseline algorithms.
  
  \item We curate a benchmark dataset, the Generative Model Zoo, which includes a diverse compilation of more than 250 community-created generative models, 1000 single-image fine-tuned models, and a set of ground truth (query, model) pairs for evaluating the retrieval algorithm on different query modalities.
\end{itemize}

\section{Related Works}
\lblsec{related}

\myparagraph{Deep generative models.}
Generative models are open-sourced at a rate of thousands per month. They use different learning objectives~\cite{kingma2013auto,goodfellow2014generative,oord2016conditional,ho2020denoising,song2020score}, training techniques~\cite{karras2020ADA,mokady2022self,Sauer2021NEURIPS,rombach2021highresolution}, and network architectures~\cite{razavi2019generating,brock2019large,karras2019style,esser2021taming}. They are also trained on different datasets~\cite{mokady2022self,schultz2020freagan,yu2015lsun,choi2020starganv2} for different applications~\cite{ha2018world,zhang2021datasetgan,zhu2021barbershop,lewis2021tryongan,chen2020deepfacedrawing,albahar2021pose,patashnik2021styleclip}. %
This diversity leads to the question, among all the models, which one shall we use? 
Our goal is \emph{not} to introduce a new model. Instead, we want to help researchers, students, and artists find existing models more easily. 

\begin{figure*}[!]
    \centering
    
    \includegraphics[width=\linewidth]{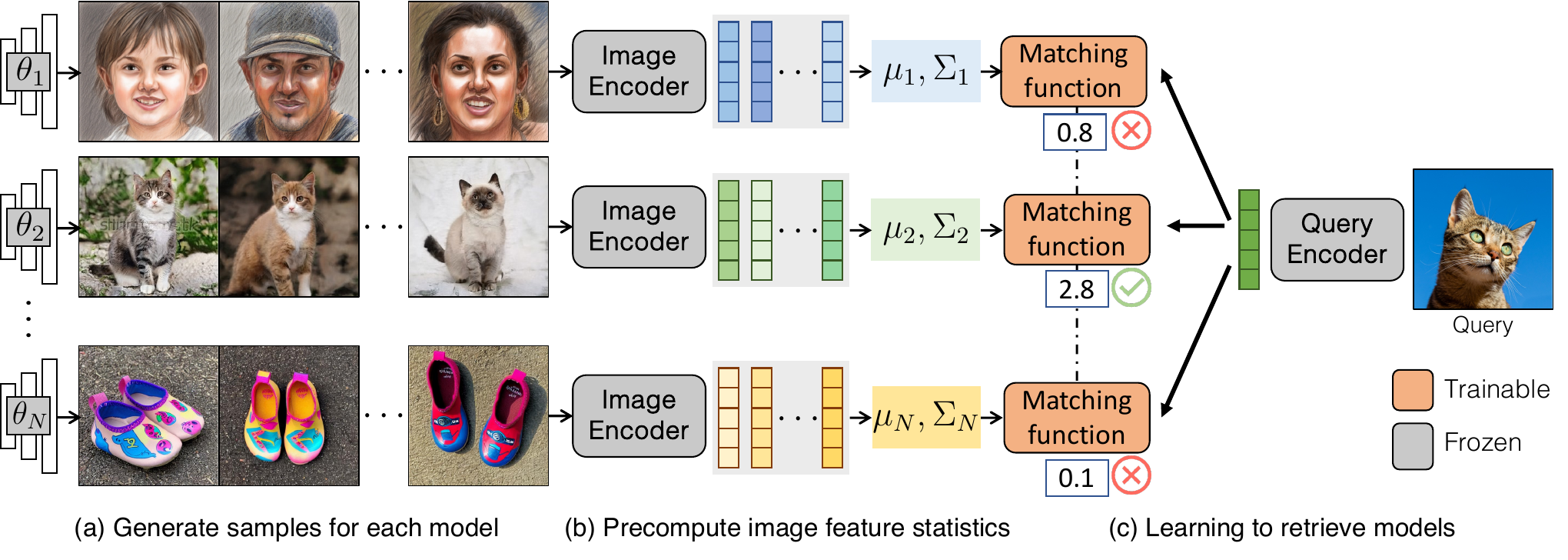}
    \caption{\textbf{Overview.}
    Given a collection of models $\gcolle$, (a) we first generate samples for each model $\gi$, and (b) encode the samples into features and compute the $1^{\text{st}}$ and $2^{\text{nd}}$ order feature statistics for each model. The statistics are cached for efficiency. (c) We then learn a matching function using contrastive loss that takes as input the query feature and model statistics and returns the similarity score between the model and query. The models with the best similarity scores are retrieved. We support queries of different modalities (text, image, or sketch). Photo by \href{https://unsplash.com/@cedric_photography}{@cedric\_photography} (right).}
    \lblfig{method}
\end{figure*}

\myparagraph{Model editing and fine-tuning}
\sysubmit{
As methods for editing and fine-tuning generative models become more accessible and efficient, these have also contributed to a proliferation of models.
Several works edit a pre-trained generative model with simple user interfaces like sketching~\cite{wang2021sketch}, warping~\cite{wang2022rewriting}, blending~\cite{bau2020rewriting}, and text prompts~\cite{gal2021stylegan}. Besides editing a model, a line of works propose methods to fine-tune models to match a small collection of images~\cite{wang2018transferring,karras2020ADA,zhao2020diffaugment,mo2020freeze,noguchi2019image,zhao2020leveraging,li2020few,ojha2021few-shot-gan,wang2020minegan,nitzan2022mystyle}. Various works further improve the fine-tuning speed~\cite{liu2020towards,Sauer2021NEURIPS,kumari2021ensembling,grigoryev2022and}.}

\myparagraph{Customization of large-scale diffusion models.}
Recently, large-scale text-to-image models~\cite{rombach2021highresolution,ramesh2022hierarchical,saharia2022photorealistic} have shown exemplary performance in generating diverse styles and compositions given the text prompt, with downstream versatile image editing capacity~\cite{avrahami2022blended,hertz2022prompt,tumanyan2022plug,zhang2023adding,brooks2022instructpix2pix}. But these models cannot still synthesize personalized and artistic concepts that are hard to describe via text. To enable that, various works have proposed model customization techniques~\cite{ruiz2022dreambooth,gal2022image,kumari2022multi,kawar2022imagic,han2023svdiff}. This has given rise to tens of thousands of fine-tuned models for different styles and concepts~\cite{huggingfacehub,civitailink}, thus making model search increasingly relevant. Recently, several works use retrieval-augmented generative models to improve the fidelity of rare entities~\cite{blattmann2022retrieval,chen2022re,ma2023unified,casanova2021instance}. Different from these, our goal is to simply find a model relevant to user queries, rather than generating an exact image given the query.

\myparagraph{Content-based retrieval.}  Building upon classical information retrieval~\cite{baeza1999modern, manning2010introduction}, content-based retrieval deals with queries over image, video, or other media~\cite{gudivada1995content,datta2008image,hu2011survey}. Content-based image retrieval methods use robust visual descriptors~\cite{oliva2001modeling,lowe2004distinctive,dalal2005histograms} to match objects within video or images~\cite{sivic2003video,arandjelovic2012three}. Methods have been developed to compress visual features to scale retrieval~\cite{torralba2008small,weiss2008spectral,jegou2010aggregating,gong2012iterative}, and deep learning has enabled compact vector representations for retrieval~\cite{torralba2008small,krizhevsky2011using,babenko2014neural,zheng2017sift}. 
In addition to image queries, various works have proposed methods for sketch-based retrieval~\cite{eitz2010sketch,lin20133d,yu2016sketch,sangkloy2016sketchy,liu2017deep,radenovic2018deep,ribeiro2020sketchformer}. There is a growing interest in joint visual-language embeddings~\cite{frome2013devise,socher2014grounded,karpathy2014deep,faghri2017vse,jia2021scaling,radford2021learning} that enable text queries for image content.  We also adopt deep image representations for our setting, but unlike single-image retrieval, we index \textit{distributions} of images that cannot be fully materialized. Concurrent to our work, HuggingGPT~\cite{shen2023hugginggpt} also explores model retrieval but for user-defined tasks such as object detection.

\section{Methods}
\lblsec{method}

In this section, we develop a retrieval framework for deep generative models. When a user specifies an image, sketch, or text query, we would like to retrieve a model that best matches the query. %
We denote a collection of $N$ models by $\{\theta_1,\theta_2,\dots,\theta_N\}$ and the user query by $\qry$, and we assume a uniform prior over models (i.e., $\gcolle$). Every model $\gi$ captures a distribution $p ( x|\theta ) $ over images $\im$. While prior retrieval methods~\cite{smeulders2000content,manning2010introduction} search for single instances, we aim to construct a method for retrieving distributions.

To achieve this, we introduce a probabilistic formulation for generative model retrieval. \reffig{method} shows an overview of our approach. Our formulation is general to different query modalities and various types of generative models, and can be extended to different algorithms. In \refsec{probabilistic}, we derive our model retrieval formulation based on a Maximum Likelihood Estimation (MLE) objective based on pre-trained deep learning features. In \refsec{learn}, we further propose a contrastive learning method to adapt features to different query modalities and model search task. 
We present our model retrieval algorithms for an image, a text, and a sketch query, respectively. In Sections~\ref{sec:zoo} and \ref{sec:method_app}, we discuss our new benchmark and user interface for model search. 

\subsection{Probabilistic Retrieval for Generative Models}
\lblsec{probabilistic}
Our goal is to quantify the posterior probability of each model $\gi$ given the user query $\qry$, and retrieve the model with the maximum $\mdlprob$. We define the probabilistic model retrieval objective as:
\begin{equation}
    \max_{\gset} \mdlprob = \frac{p(\qry | \g) p(\g)}{p(\qry)} \propto \max_{\gset} p(\qry | \g).
    \lbleq{intro}
\end{equation}

As we assume models are uniformly distributed, it is equivalent to finding the likelihood of the query under each model, $p(\qry | \g)$.

There are two scenarios for inferring the query likelihood $p(\qry | \g)$: (1) When the query  $\qry$ shares the same modality with the model  $\g$ (e.g., searching models with image queries), we directly reduce the problem to estimating the model's density, (2) When the query $\qry$ has a different modality from the model $\g$ (e.g., searching models with text queries), we use cross-modal similarity to estimate $p(\qry | \g)$. We discuss both cases in the following.

\myparagraph{Image-based model retrieval.} 
\label{sec:ibmr}
Given an image query $\qry$, we directly estimate the likelihood of the query $p(\qry | \g)$ from each model. In other words, we seek to find the model that is most likely to generate the query image. Since density estimation is intractable for many generative models (e.g., GANs~\cite{goodfellow2014generative}, VAEs~\cite{kingma2013auto}), we approximate each model as a Gaussian distribution over image features~\cite{heusel2017gans}. After we sample an image from each model, denoted by $\im$, we obtain its image feature $\imfeat \coloneqq \phi_{im}(\im)$, where $\phi_{im}$ is a frozen feature extractor. We now express $p(\qry | \g)$ in terms of image features $\imfeat$, so \refeq{intro} becomes:
\begin{equation}
    \centering
    \begin{aligned}
    &\max_{\gsett} %
    (2\pi|\Sigma_\mi|)^{-\frac12}
    \exp\left(-\frac12(\imfeat_{\qry} - \mu_\mi)^T\Sigma_\mi^{-1}(\imfeat_{\qry} - \mu_\mi)\right),
    \end{aligned}
\end{equation}

\noindent where the query image feature is denoted by $\imfeat_{\qry} \coloneqq \phi_{im}(\qry)$, and each model $\gi_n$ is approximated by $p(\imfeat|\gi_n)\sim\mathcal{N}(\mu_\mi, \Sigma_\mi)$. We refer to this method as \texttt{\gauden}. %

\myparagraph{Text-based model retrieval.}
\label{sec:tbmr}
Given a text query $\qry$ and a generative model $p(\im | \g)$ capturing a distribution of images $\im$, we want to estimate the conditional probability $p(\qry | \g)$.
\begin{equation}
\begin{aligned}
    p(\qry | \g) = \int p(\qry | \im) p(\im | \g) d\im \propto \int \frac{p(\im | \qry)}{p(\im)} p(\im | \g) d\im.
\end{aligned}
\label{eq:general_eq}
\end{equation}
Here we assume conditional independence between query $\qry$ and model $\g$ given image $x$, so $p(\qry | \im, \g) = p(\qry | \im)$, and we apply Bayes' rule to get the final expression. 
In \refeq{general_eq}, a text query $\qry$ may correspond to multiple possible image matches $p(\im|\qry)$, so the expression cannot be simplified the same way as image-based model retrieval. Instead, we can view \refeq{general_eq} as an integral of the mutual information between $\qry$ and $\im$, which can be estimated with a contrastive representation~\cite{oord2018representation}. Thus, we use CLIP similarity~\cite{radford2021learning} to approximate $\frac{p(\im | \qry)}{p(\im)} \propto \exp(\frac{1}{\tau}\nmimclip^T\nmtxtclip)$, where $\nmimclip \coloneqq \clipimex(x)$ and $\nmtxtclip \coloneqq \cliptxtex(\qry)$ are the normalized CLIP image and text features; $\tau$ is a temperature term.

To approximate the integration, one can sample images $\im$ from $p(\im | \g)$ and then take the average of the CLIP similarities between images and query. We refer to this method as \texttt{\mc}. 
To further speed up computation, we can pre-compute the mean of CLIP image embeddings for each model, and at inference time, we directly evaluate similarities between the query embedding and the mean embedding, as follows:

\begin{equation}
    \centering
    \begin{aligned}
    &\int \frac{p(\im | \qry)}{p(\im)} p(\im | \g) ~ d\im \approx \exp\left(\frac{\Tilde{\mu_\mi}^T\nmtxtclip}{\tau}\right) , \\
    \text{where}\;\; \mu_\mi =  &\;\mathbb{E}_{\nmimclip \sim p(\nmimclip | \gi_n)} \left[ \nmimclip \right] ; \;  \Tilde{\mu_\mi} = \frac{\mu_\mi}{||\mu_\mi||} .
    \end{aligned}
    \label{eq:first_mom}
\end{equation}
We find that this approximation works well in practice, and we refer to this method as \texttt{\firstmom}. Derivation details for \texttt{\mc} and \texttt{\firstmom} can be found in \refapp{derive}.

\myparagraph{\firstmom\  method for image queries.}
Likewise, we find that applying \texttt{\firstmom} for image-based model retrieval yields a performance close to  \texttt{\gauden}. Interestingly, for sketch queries, \texttt{\firstmom} method outperforms \texttt{\gauden}. While CLIP shows strong cross-modal matching performance between sketch queries and image models, the domain gap makes density estimation a less favorable option for sketch queries compared to the \texttt{\firstmom} method. We provide more discussions in \refsec{expr}.

\subsection{Learning to Retrieve Models}
\lblsec{learn}
We have outlined several ways to approximate $p(\theta|q)$ in \refsec{ibmr} to perform content-based model retrieval. However, these approximations involve only pre-trained feature extractors and may not be optimal for model retrieval with certain query modalites. For example, the frozen features that work well for images may struggle with sketches. To address this issue, we introduce a set of learnable parameters to fine-tine the approximations of $p(\theta|q)$ on our Generative Model Zoo dataset (\refsec{zoo}).

Under the contrastive learning framework, we maximize\\$\mathbb{E}\left[\log \frac{p(\theta|q)}{p(\theta)}\right]$, the mutual information between the query $q$ and the model $\theta$. Since we assume a uniform prior over the models, the mutual information is effectively our retrieval objective $\mathbb{E}\left[\log p(\theta|q)\right]$. The training dataset $\mathcal{D}$ consists of sample statistics for $M$ models and a number of ground-truth query-model pairs. We optimize a matching function $s_\psi$ that yields a query-to-model similarity score, by minimizing the InfoNCE loss~\cite{oord2018representation}:
\begin{gather}
     \min_{\psi} \mathop{\mathbb{E}}_{(q,\theta_i)\sim \mathcal{D}}\Bigg[-\log \frac{\exp(s_\psi(q, \theta_i)/\tau) }{\mathop{\sum}_{j=1}^{M} \exp(s_\psi(q, \theta_j)/\tau)} \Bigg], \\
     s_\psi(q, \theta_i) = \begin{cases}
      \frac{(A_\psi \Tilde{\mu_i})^T (A_\psi \Tilde{h_q})}{||A_\psi \Tilde{\mu_i}|| ||A_\psi \Tilde{h_q}||} & \textnormal {  if \firstmom}\\
      \log \mathcal{N}(\nmtxtclip| \mu_i, ~ A_\psi \Sigma_i A_\psi^T) & \textnormal {  if \texttt{\gauden}}
    \end{cases}   
    \lbleq{learning_to_rank}
\end{gather}
where $A_\psi$ is a matrix parameterized by $\psi$,  $(\mu_i$, $\Sigma_i)$ are the feature-space sample mean and covariance of model $\theta_i$, and $\nmtxtclip$ is as calculated in the previous section using a pre-trained feature extractor. The contrastive learning objective maximizes the similarity score between query $q$ and the ground truth model and minimizes the similarity between the query and other models.

We define the matching function based on two formulations described in \refsec{ibmr}: \texttt{\firstmom} and \texttt{\gauden}. For \texttt{\firstmom}, we project the original feature using the matrix $A_\psi$. For \texttt{\gauden}, $\mathcal{N}(\nmtxtclip| \mu_i, ~ A_\psi \Sigma_i A_\psi^T)$ denotes the gaussian PDF with parameters $\{\mu_i, ~ A_\psi \Sigma_i A_\psi^T\}$, evaluated at the query feature~$\nmtxtclip$. We experiment with different $A_\psi$ parameterizations, including full, triangular, and diagonal matrices. For the diagonal case, we limit the range of $A_\psi$ such that $A_\psi = \text{diag}(\text{sigmoid}(\psi))$

In~\refsec{expr}, we show that our method can retrieve models that share similar visual concepts with the query. %

\begin{figure}
    \centering
    \includegraphics[width=\linewidth]{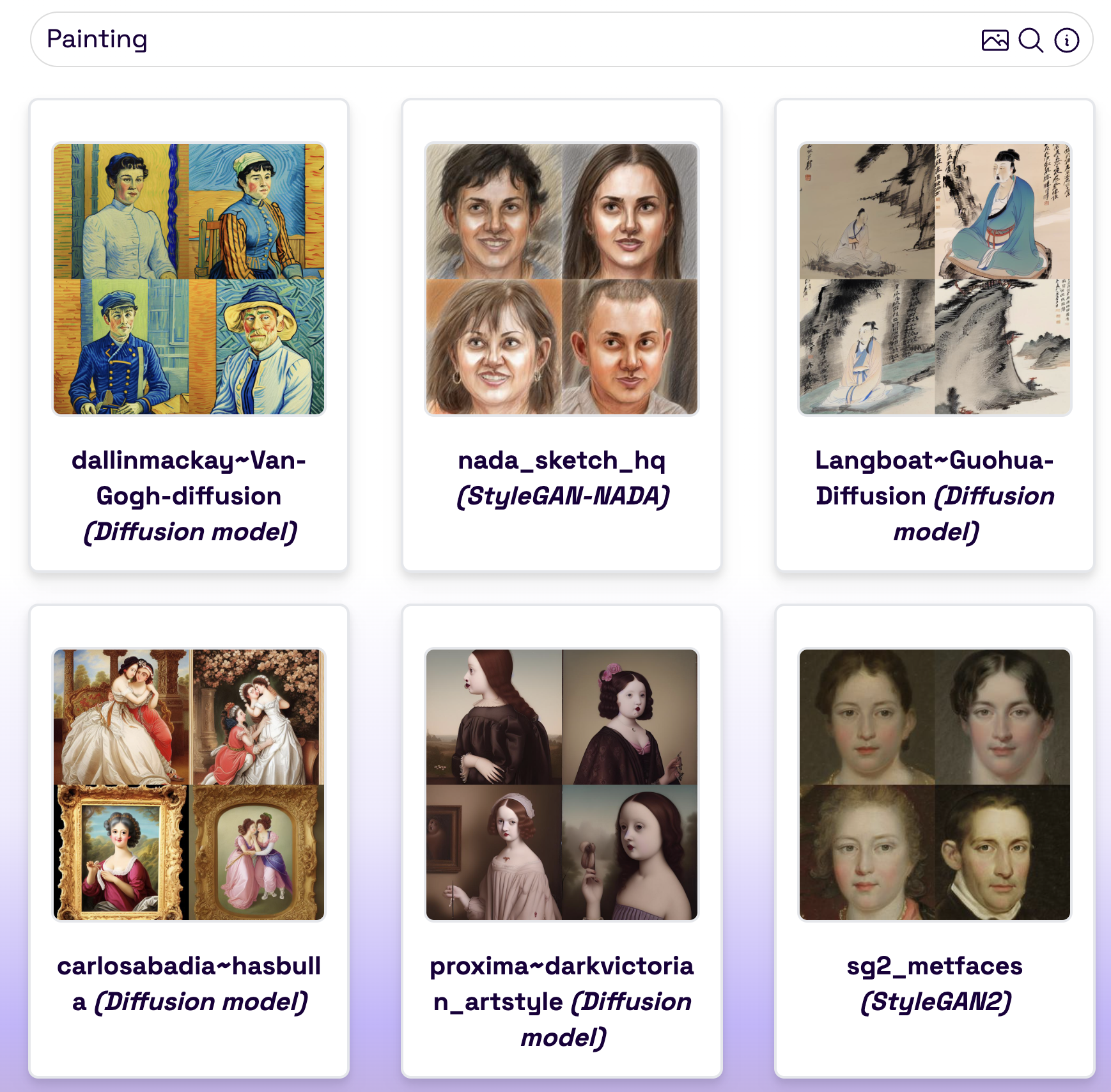}
    \caption{\textbf{User interface.} In our \href{https://modelverse.cs.cmu.edu/}{model search platform}, a user can submit a text query, an image query, or a combination of both to retrieve generative models that best match the query.
    Here we show the top 6 retrievals for the text query ``painting''. %
     The user can further explore additional samples from the models or search for models similar to a particular model.}%
    \lblfig{interface}
\end{figure}

\section{The Generative Model Zoo}
\lblsec{zoo}

We introduce a new benchmark, the Generative Model Zoo, consisting of a set of generative models capturing a variety of architectures and subject areas for evaluating retrieval performance.

\myparagraph{Internet model zoo}
It consists of a total of 259 publicly available generative models trained using different techniques, including GANs~\cite{karras2020analyzing,Karras2021alias,karras2018progressive,mokady2022self,kumari2021ensembling,gal2021stylegan,wang2021sketch,wang2022rewriting,sauer2022stylegan,pinkney2020awesome,lucidlayersstylegan3}, diffusion models~\cite{ho2020denoising,song2020denoising,dhariwal2021diffusion,ruiz2022dreambooth}, MLP-based generative model CIPS~\cite{anokhin2020image}, and the autoregressive model VQGAN~\cite{esser2021taming}.
Out of the 259 models, 23 were created via fine-tuning by individual users, 133 were obtained from the GitHub repositories of academic papers, and the remaining models were collected from public model sharing website~\cite{huggingfacehub}. This includes models like DreamBooth~\cite{ruiz2022dreambooth} that are finetuned to generate a specific concept (e.g., a unique toy) with a fixed text prompt as input. Hence, to index these models, we generate samples conditioned on the unique text prompt used to fine-tune the model. A comprehensive list of the models and their respective sources is included in \refapp{details}.

We create a set of benchmark queries for each model in multiple modalities, i.e., text, image, and sketch. The text queries include human-written textual queries that describe the images in a model. Image queries are created by sampling images from models, and sketch queries are created by using the method of Chan \emph{et al.}~\shortcite{chan2022learning}. %

\myparagraph{Synthetic model zoo.}
To further test our method at scale, we create a synthetic model zoo of 1,000 models. Each model is finetuned from the pretrained Stable Diffusion v1.4 checkpoint~\cite{rombach2021highresolution} on a single image. The models are fine-tuned on instances from animal classes in ImageNet~\cite{deng2009imagenet} or artistic images downloaded from Unsplash~\cite{Unsplash}. We manually pick 50 ImageNet classes and 10 instance images from each class, thus training 500 models with instance images from the ImageNet Dataset. For the artistic models, we select 500 images from Unsplash that match the keyword ``art''. To fine-tune the model, we randomly select between Dreambooth~\cite{ruiz2022dreambooth} with LoRA~\cite{hu2021lora} and Custom Diffusion~\cite{kumari2022multi}. In this case, we only create image and sketch queries, as describing specific instantiation of a general category from ImageNet classes is difficult via only text. Similarly, describing an art work using text is also non-trivial.

\section{User interface}\label{sec:method_app}

Our baseline method is fast enough for interactive use, and we create a web-based UI for the search algorithm. The user can search for models by entering text or uploading an image/sketch. %
The interface displays models that best match the query, where clicking on a model's thumbnail shows more image samples. The website utilizes a backend GPU server to enable real-time model search on any client device. \reffig{interface} shows a screenshot of our UI. %

\section{Experiments}
\lblsec{expr}
For our baseline retrieval methods, we benchmark their performance over text, image, and sketch modalities and discuss several algorithmic design choices.

\myparagraph{Implementation details.} 
\dinorm{We test our method with the commonly used pretrained feature extractor} -- CLIP~\cite{radford2021learning}. %
We use the $\ell_2$-normalized CLIP features, as discussed in~\refsec{probabilistic}. 

We pre-calculate each model's generated distribution mean and covariance in the CLIP feature space, following the pre-processing steps described in \texttt{clean-fid}~\cite{parmar2021cleanfid}. For ImageNet fine-tuned models, we generate class-specific prompts used to sample images using ChatGPT. For artistic fine-tuned models, we use general prompts, e.g., ``a painting in the style of <specific art style>''. More implementation details are in \refapp{details}.

To evaluate model search, we use the top-k accuracy metric, i.e., frequency of finding the \textit{unique} ground truth model within the top-k retrieved models using the query corresponding to that model.

\subsection{Model Retrieval}

\begin{table}[!t]
\caption{\textbf{Image- and sketch-based model retrieval}. We compare retrieval performance with pre-trained and fine-tuned scoring functions utilizing \dinorm{CLIP network backbone}. We observe the best performance with CLIP and fine-tuning. The (FT) suffix denotes the learning-based method (\refeq{learning_to_rank}) using the optimal parameterization selected via first doing cross-validation on the \camready{\emph{Synthetic Model Zoo}} as shown in \reftbl{cross_validation}.
}\lbltbl{image_retrieval_results}
\resizebox{\linewidth}{!}{
\begin{tabular}{llrrr}
\toprule
      &     & \multicolumn{3}{c}{Top-k Accuracy}  \\
      \cmidrule(lr){3-5}
      &     & Top-1     & Top-5     & Top-10     \\ 
      \midrule \midrule
\multirow{8}{*}{\rotatebox[origin=c]{90}{\textbf{Image (Gen.)}}}
& CLIP+Monte-Carlo (2.4K)      &0.82&	\textbf{0.96} &	\textbf{0.99} \\
& CLIP+best-k (2.4K, k=1)     &0.80 & 0.95 & \textbf{0.99} \\
& CLIP+best-k (2.4K, k=10)     &0.82 & \textbf{0.96} & \textbf{0.99} \\
\cmidrule(l{6pt}r{6pt}){2-5}
& CLIP+\firstmom        & 0.78 &	0.94 &	0.98  \\
& CLIP+\firstmom~(FT)        & 0.81 &	0.95 &	0.98  \\
& CLIP+\gauden        & 0.84 &	\textbf{0.96} &	0.98 \\
& CLIP+\gauden~(FT)        & \textbf{0.85} &	\textbf{0.96}&	\textbf{0.99} \\
      \midrule
\multirow{8}{*}{\rotatebox{90}{ \begin{tabular}{@{}c@{}} \textbf{Sketch}\end{tabular} }}
& CLIP+Monte-Carlo (2.4K)     &0.32 & 0.59 & 0.72 \\
& CLIP+best-k (2.4K, k=1)     &0.28 & 0.55 & 0.68 \\
& CLIP+best-k (2.4K, k=10)     &0.32 & 0.59 & 0.72 \\
\cmidrule(l{6pt}r{6pt}){2-5}
& CLIP+\firstmom        &0.32 &	0.61&	0.73 \\
& CLIP+\firstmom~(FT)        & \textbf{0.49} &	\textbf{0.75} &	\textbf{0.84}  \\
& CLIP+\gauden        &0.18	& 0.39 &	0.53  \\
& CLIP+\gauden~(FT)        & 0.30 &	0.56 &	0.69 \\
\bottomrule
\end{tabular}}

\end{table}
\myparagraph{Model retrieval via image and sketch queries.} We report top-k accuracy of model retrieval on the \emph{Internet model zoo} dataset in \reftbl{image_retrieval_results}. We show the performance of different formulations with \dinorm{CLIP features}. %
We train the transformation matrix $A_{\psi}$ in the learning to retrieve method using the \emph{Synthetic model zoo}. The retrieval performance is best with CLIP features combined with our learning \dinorm{method, especially for sketch-based retrieval.} The learning method is denoted with the suffix (FT) in \reftbl{image_retrieval_results}. For comparison, we also include a modified k-NN algorithm in which we find the query's feature-space k-nearest neighbors among each model's image samples and sort the models by the mean distance to those k samples. We refer to this baseline as ``best-k neighbors" or simply ``best-k". \reffig{results_main} shows qualitative examples of model retrieval for different image and sketch queries.

\begin{table}[!t]
\caption{\textbf{Selecting best parameterization for $A_\psi$ with cross-validation.} We fine-tune different versions of similarity function and transformation matrix parameterization with 5-fold cross-validation on the \emph{Synthetic Model Zoo}. We use it to determine the optimal parameterization of $A_\psi$ (marked with a ``$\star$'') for each method and modality.}
\resizebox{\linewidth}{!}{
\begin{tabular}{llrrcc}
\toprule
      &     &  & & \multicolumn{2}{c}{Mean Testing Accuracy}  \\
      \cmidrule(lr){5-6}
      &     & Modality & $A\psi$  & \shortstack[c]{Top-1\\ (FT)}     &  \shortstack[c]{Top-1\\ (Pre-trained)} \\ 
      \midrule \midrule
\multirow{12}{*}{\rotatebox[origin=c]{90}{ \begin{tabular}{@{}c@{}} \textbf{Synthetic Zoo}\end{tabular} }}
& CLIP+\firstmom        & Image & Diagonal & 0.84 &	0.81  \\
& CLIP+\firstmom        & Image & $\star$Triangular & \textbf{0.90}  &	0.81  \\
& CLIP+\firstmom        & Image & Full & 0.89  &	0.81  \\
& CLIP+\gauden        & Image & Diagonal & \textbf{0.91}	&	0.88 \\
& CLIP+\gauden        & Image & $\star$Triangular & \textbf{0.91}	&	0.88 \\
& CLIP+\gauden        & Image & Full & \camready{\textbf{0.91}}	&	0.88 \\
\cmidrule{2-6}
& CLIP+\firstmom        & Sketch & Diagonal &	0.30 &	0.15  \\
& CLIP+\firstmom        & Sketch & $\star$Triangular & \textbf{0.61} &	0.15  \\
& CLIP+\firstmom        & Sketch & Full & \textbf{0.61}  &	0.15  \\
& CLIP+\gauden        & Sketch & Diagonal &	0.29 &	0.09 \\
& CLIP+\gauden        & Sketch & $\star$Triangular & \textbf{0.51}	&	0.09 \\
& CLIP+\gauden        & Sketch & Full &	 0.38 &	0.09 \\
\bottomrule
\end{tabular}}
\lbltbl{cross_validation}

\end{table}

\looseness=-1 \myparagraph{Learning to retrieve models.}
We conduct 5-fold cross-validation on \emph{Synthetic Model Zoo} to determine the best parameterization of the transformation matrix $A_\psi$ for the learnable matching function (\refeq{learning_to_rank}). \reftbl{cross_validation} shows the test-set top-1 accuracy of each method compared to the baseline, which uses the pre-trained feature extractor. We find that our contrastive learning method consistently improves the pretrained features for various query modalities and formulations. We select the best parameterization via cross-validation for training the matching function from scratch on the full \emph{Synthetic Model Zoo} and then test on the \emph{Internet Model Zoo}.  The learned matching function generalizes to the \emph{Internet Model Zoo} as shown in \reftbl{image_retrieval_results}. We also train our method on \emph{Internet Model Zoo} and qualitatively test its generalization on \emph{Synthetic Model Zoo} in \reffig{result_additional_synth}. Please refer to \refapp{analysis} for a quantitative analysis.

\begin{table}[!t]
\caption{\textbf{Learning to retrieve with text queries}. We show the result of the learning-based method with text queries using a 5-fold cross-validation on the \emph{Internet Model Zoo} itself. This is because we have ground truth text queries only for that, as discussed in \refsec{zoo}. We optimize different parameterizations of the transformation matrix on the training split and show retrieval performance on the testing split of 51 models.}
\resizebox{\linewidth}{!}{
\begin{tabular}{llrrcc}
\toprule
      &     &  & & \multicolumn{2}{c}{Mean Testing Accuracy}  \\
      \cmidrule(lr){5-6}
      &     & Modality & $A\psi$   & \shortstack[c]{Top-1\\ (FT)}     &  \shortstack[c]{Top-1\\ (Pre-trained)} \\ 
      \midrule \midrule
\multirow{6}{*}{\rotatebox[origin=c]{90}{\shortstack[c]{\textbf{Internet}\\ \textbf{Zoo} }}}
& CLIP+Monte-Carlo (2.4K)        & Text & - &	- &	0.75  \\
& CLIP+best-k (2.4K, k=1)        & Text & - &	- &	0.67  \\
& CLIP+best-k (2.4K, k=10)        & Text & - &	- &	0.74  \\
\cmidrule(l{6pt}r{6pt}){2-6}
& CLIP+\firstmom        & Text & Diagonal &	0.78 &	0.77  \\
& CLIP+\firstmom        & Text & Triangular & 0.80 &	0.77  \\
& CLIP+\firstmom        & Text & Full &	\textbf{0.81} &	0.77  \\
\bottomrule
\end{tabular}}
\lbltbl{cross_validation_text}. 
\end{table}

\myparagraph{Text-based model retrieval.} 
We show the results of text-based retrieval on \emph{Internet Model Zoo}, i.e., collected publicly available generative models, which we manually labeled with corresponding text descriptions. 
We use the CLIP~\cite{radford2021learning} as the pretrained feature extractor since CLIP has both text and image encoders. Since our learning-based method requires a training and test split, we perform 5-fold cross-validation and show the mean Top-1 accuracy for all methods, including baselines, to be consistent. \reftbl{cross_validation_text} shows the retrieval performance of both methods. Similar to image and sketch-based retrieval, we also include results for the best-k neighbors baseline. The proposed baseline method achieves an accuracy of $77\%$, while the best learning-based method outperforms it with $81\%$ accuracy.

\reffig{results_main} shows qualitative examples of the top three and lowest-ranked retrieval given a text query with the $\firstmom$ method . %
Both quantitative numbers and visual inspection of results show that our method retrieves relevant generative models. We also analyze the retrieval score of all models corresponding to a given query. For object categories, such as ``dogs'' or ``buses'', we observe a clear drop in retrieval score for irrelevant models. For broader queries, such as ``indoors'', ``modern art'' and ``painting'', the drop-off is gradual. We show detailed analysis in \refapp{analysis}.

\myparagraph{Comparison with metadata-based search.} 
We compare against an alternative where we index models using user-defined descriptions and search for models by text-matching the query and the model descriptions.
To test this, we use the text query associated with each model as its description (e.g., ``portraits with Botero's style'' for one of the StyleGAN-NADA models). 

\sysubmit{We test metadata-based search on image- and sketch-based retrieval, with two ways of description-matching. (1) \texttt{Description (text)}: we caption image/sketch queries using BLIP~\cite{li2022blip} and retrieve models whose descriptions contain any nouns, verbs, or adjectives (i.e., any non-filler words) that appear in the caption.
(2) \texttt{Description (CLIP)}: we embed model descriptions into CLIP features and compare the similarity between the image/sketch query feature and the description feature.}

As shown in~\reftbl{clip_tag_retrieval_results}, the content-based search methods outperform the baselines while they also obviate the need for model descriptions. Metadata-based search typically fails when captions are incorrect or when the object-centric tags cannot describe the model fully. Creating comprehensive descriptions might reduce the gap, but it is time-consuming to describe every visual aspect and anticipate users’ queries in advance. Our evaluation does not include text-based queries, as we use the model descriptions as the ground truth. However, we expect the baselines to have the same limitation for text queries in the case where a text query requests a visual concept that the tags fail to enumerate.

\looseness=-1 \myparagraph{Running time and memory.} %
Time and memory efficiency are crucial for supporting many concurrent user searches over %
a large-scale model collection. 
\sysubmit{While \texttt{\firstmom} obtains competitive top-k accuracies, it runs 3.2-7.5 $\times$ faster than \texttt{\mc} or \texttt{\gauden}. The fine-tuned \texttt{\firstmom} method further improves accuracy, but runs at speeds more on par with \texttt{\mc} or \texttt{\gauden}.  \texttt{\firstmom} is extremely memory-efficient. \suppmat{Further analysis is in \refapp{analysis}.}}

\section{Extensions}
\lblsec{applications}
We describe two extensions:  searching with multimodal queries and searching with a given model. 

\myparagraph{Multimodal user query.}
We show qualitatively that our search method can be extended to multimodal queries based on the Product-of-Experts formulation~\cite{hinton2002training, huang2021multimodal}. Given a multimodal query (e.g., text-image pair), the retrieval score is the product of individual query likelihoods. We demonstrate how one can leverage multiple input modalities to perform nuanced searches in \reffig{multimodal} using the \texttt{\firstmom} method with pre-trained features. We show additional multimodal retrieval results in \refapp{analysis}.

\myparagraph{Generative model query.}
\sysubmit{Given the generative model collection, we can also retrieve similar models based on the cosine similarity between the feature-space means of query models and other generative models.} 
\reffig{similar_model_retrieval} shows qualitative examples of similar-model retrieval using CLIP feature space.

\begin{table}[!t]
\caption{\textbf{Comparison with metadata-based search.} We compare our \texttt{\gauden} content-based search method with two metadata-based search methods: (1) \texttt{Description (text)}: we caption image/sketch queries using BLIP~\cite{li2022blip} and select models whose descriptions contains any non-filler words from the caption, (2) \texttt{Description (CLIP)}, we embed model descriptions into the CLIP feature space and compare the similarity between the CLIP query features and the CLIP description features.}
\resizebox{\linewidth}{!}{
\begin{tabular}{llrrr}
\toprule
      &     & \multicolumn{3}{c}{Top-k Accuracy} \\
      \cmidrule(lr){3-5}
      &     & Top-1     & Top-5     & Top-10\\ 
      \midrule \midrule
\multirow{3}{*}{\rotatebox[origin=c]{90}{\begin{tabular}{@{}c@{}} \textbf{Image}  \\ \textbf{(Gen.)} \end{tabular}}}

& Description (text)      & 0.07 & 0.20 & 0.29 \\
& Description (CLIP)      & 0.45 & 0.72 & 0.85\\
\cmidrule(l{6pt}r{6pt}){2-5}
& CLIP+\gauden   \; (ours)    & \textbf{0.89} & \textbf{0.98} & \textbf{0.99}\\

      \midrule
\multirow{3}{*}{\rotatebox[origin=c]{90}{ \begin{tabular}{@{}c@{}c@{}} \textbf{Sketch}\end{tabular} }}
& Description (text)      & 0.07 & 0.20 & 0.28 \\
& Description (CLIP)      & 0.22 & 0.46 & 0.60 \\
\cmidrule(l{6pt}r{6pt}){2-5}
& CLIP+\firstmom   \; (ours)    &\textbf{0.49} & \textbf{0.75} & \textbf{0.84}  \\
\bottomrule     
\end{tabular}}
\lbltbl{clip_tag_retrieval_results}

\end{table}

\section{Discussion and Limitations}
\lblsec{discussion}

\sysubmit{We have introduced the new task of content-based generative model search. We have developed a data set for benchmarking model retrieval algorithms, and we have described several baselines and a contrastive learning method for further improving features.  %
}

\begin{figure}[h]
    \centering
    \includegraphics[width=\linewidth]{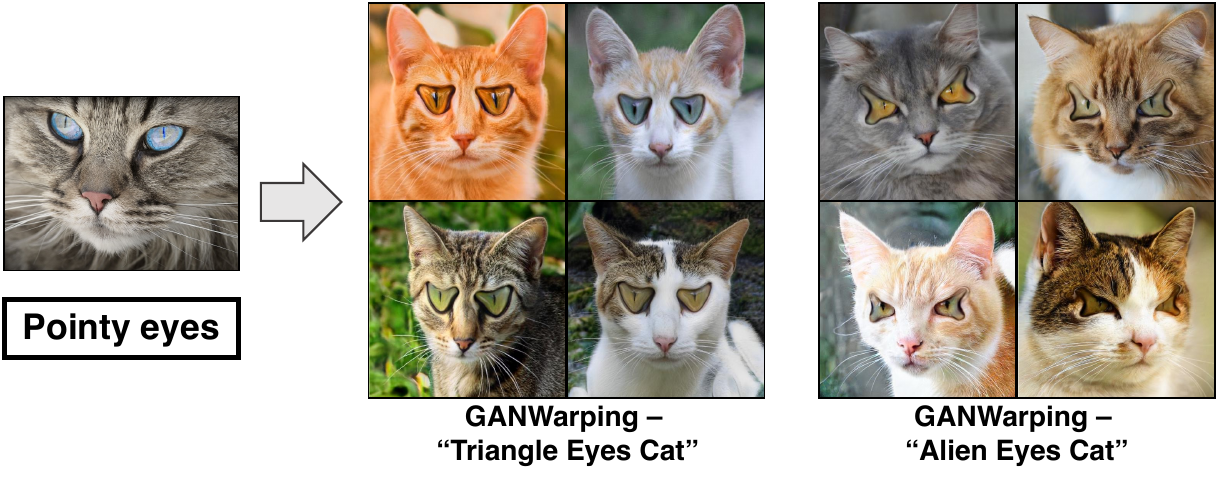}
    \caption{\textbf{Multi-modal user query.} Image and text queries combined can define nuanced concepts. When querying with the phrase ``pointy eyes'' plus a cat photo, we can retrieve ``Triangle Eyes Cat'' and ``Alien Eyes Cat''. \camready{Photo by \href{https://pixabay.com/users/cocoparisienne-127419/}{cocoparisienne} on \href{https://pixabay.com/photos/cat-tabby-face-whiskers-pet-1508613/}{Pixabay}.}}%
    \lblfig{multimodal}
\end{figure}
\begin{figure}[ht]
    \centering
    \includegraphics[width=\linewidth]{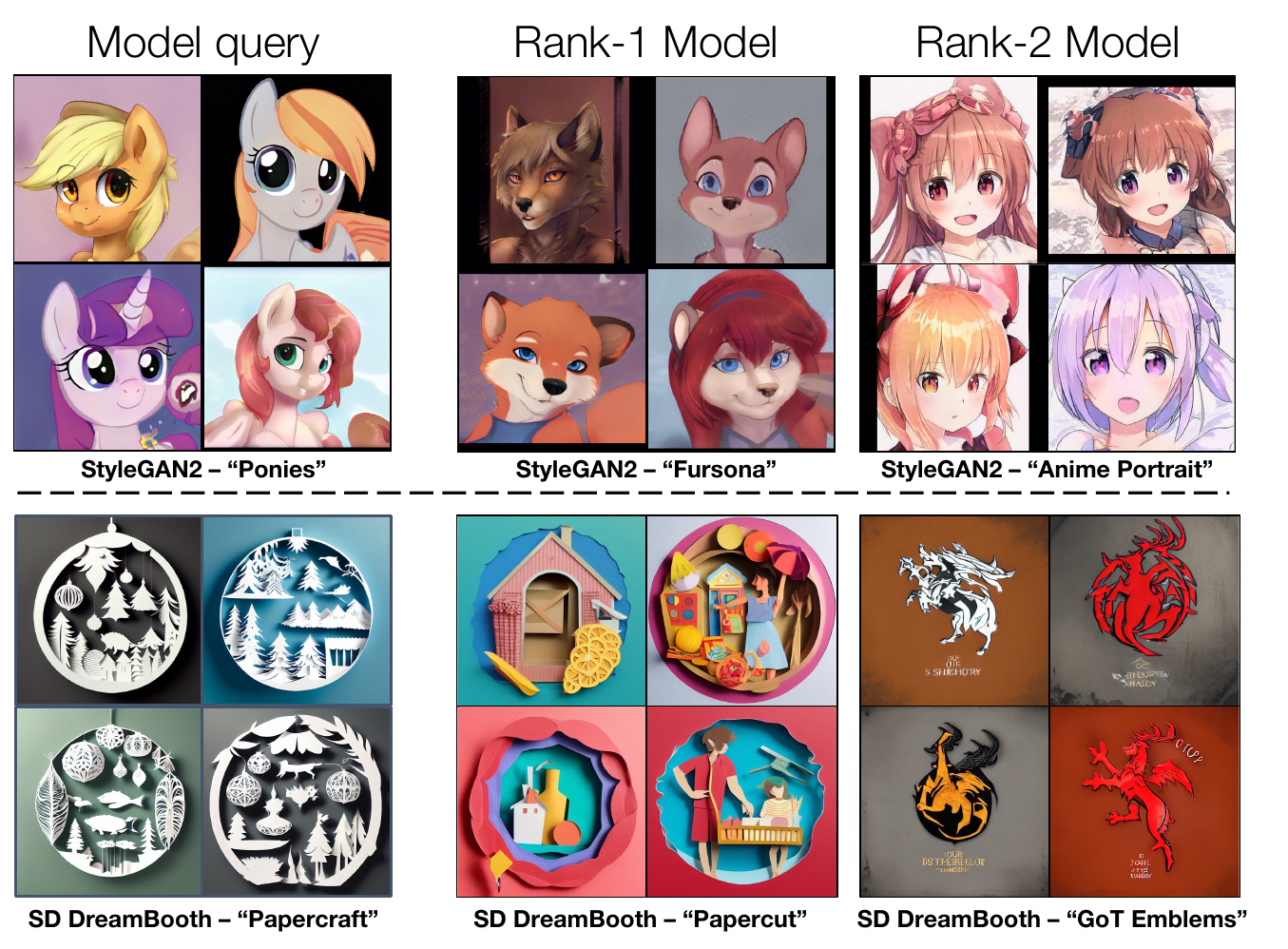}
    \caption{\textbf{Finding similar models.} Our method can find models that share similar characteristics with the query model. }
    \lblfig{similar_model_retrieval}
\end{figure}

\begin{figure}
    \centering
    \includegraphics[width=\linewidth]{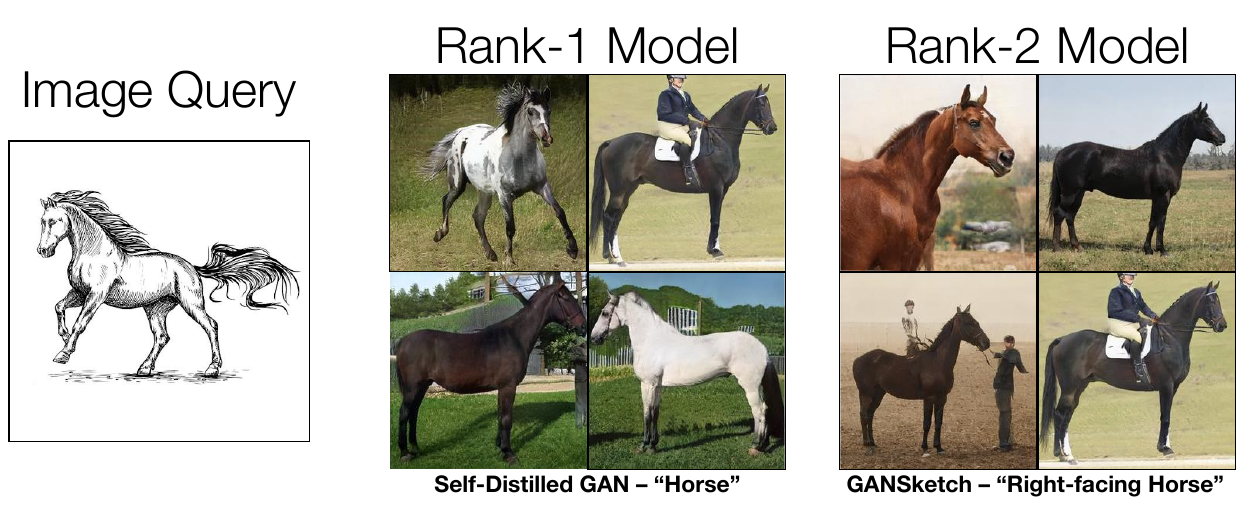}
    \caption{\textbf{Limitation.} Sometimes, a sketch query (e.g., the left-facing horse sketch) does not retrieve the most similar model at the top, which should be the GANSketch ``Left-facing Horse'' model. Sketch by \href{https://pixabay.com/users/dharmah-18954604/}{Damaris Dharmah} on \href{https://pixabay.com/}{Pixabay}.}
    \lblfig{limitation}
\end{figure}

\myparagraph{Limitations.} 
\sysubmit{While we can retrieve models in real-time based on text, image, or sketch queries, our method has several limitations. 
First, while our method is able to successfully retrieve models that capture a single concept (e.g., StyleGAN3~\cite{Karras2021alias} and DreamBooth~\cite{ruiz2022dreambooth}). There remain many future directions, including searching for personalized models that contain multiple subjects~\cite{kumari2022multi}, 3D neural objects~\cite{mildenhall2021nerf,poole2022dreamfusion}, and other media such as text, audio, and videos.

Second, our method can sometimes fail to match the user intent. A typical failure case is illustrated in \reffig{limitation}, where a query for a sketch of left facing horse retrieves general horse models instead of the most similar GANSketch ``Left-facing Horse'' model. It fails to respect the spatial orientation of the object in the query image. We show further analysis on this in \refapp{analysis}.

Nevertheless, our experiments have shown that it is realistic, possible, and useful to search an indexed collection of models by matching them against the output behavior of the models. As the number of customized and pre-trained models continues to balloon, we anticipate that search algorithms that can effectively locate relevant customized models will be an increasingly valuable resource for researchers and content creators.}

\begin{acks}
We thank George Cazenavette, Gaurav Parmar, Chonghyuk (Andrew) Song, and Or Patashnik for proofreading the draft. We are also grateful to Aaron Hertzmann, Maneesh Agrawala, Stefano Ermon, Chenlin Meng,  William Peebles, Richard Zhang, Phillip Isola, Taesung Park, Muyang Li, Kangle Deng,  and Ji Lin for helpful comments and discussion. The work is partly supported by NSF IIS-2239076, Adobe Inc., Sony Corporation, and Naver Corporation.
\end{acks}

\bibliographystyle{ACM-Reference-Format}
\bibliography{main}


\begin{thebibliography}{118}


\ifx \showCODEN    \undefined \def \showCODEN     #1{\unskip}     \fi
\ifx \showDOI      \undefined \def \showDOI       #1{#1}\fi
\ifx \showISBNx    \undefined \def \showISBNx     #1{\unskip}     \fi
\ifx \showISBNxiii \undefined \def \showISBNxiii  #1{\unskip}     \fi
\ifx \showISSN     \undefined \def \showISSN      #1{\unskip}     \fi
\ifx \showLCCN     \undefined \def \showLCCN      #1{\unskip}     \fi
\ifx \shownote     \undefined \def \shownote      #1{#1}          \fi
\ifx \showarticletitle \undefined \def \showarticletitle #1{#1}   \fi
\ifx \showURL      \undefined \def \showURL       {\relax}        \fi
\providecommand\bibfield[2]{#2}
\providecommand\bibinfo[2]{#2}
\providecommand\natexlab[1]{#1}
\providecommand\showeprint[2][]{arXiv:#2}

\bibitem[civ(2022)]%
        {civitailink}
 \bibinfo{year}{2022}\natexlab{}.
\newblock \bibinfo{title}{Civit AI}.
\newblock \bibinfo{howpublished}{\url{https://civitai.com}}.
\newblock


\bibitem[hug(2022)]%
        {huggingfacehub}
 \bibinfo{year}{2022}\natexlab{}.
\newblock \bibinfo{title}{Stable Diffusion Dreambooth Concepts Library}.
\newblock \bibinfo{howpublished}{\url{ https://huggingface.co/sd-dreambooth-library}}.
\newblock


\bibitem[Albahar et~al\mbox{.}(2021)]%
        {albahar2021pose}
\bibfield{author}{\bibinfo{person}{Badour Albahar}, \bibinfo{person}{Jingwan Lu}, \bibinfo{person}{Jimei Yang}, \bibinfo{person}{Zhixin Shu}, \bibinfo{person}{Eli Shechtman}, {and} \bibinfo{person}{Jia-Bin Huang}.} \bibinfo{year}{2021}\natexlab{}.
\newblock \showarticletitle{Pose with {S}tyle: {D}etail-Preserving Pose-Guided Image Synthesis with Conditional StyleGAN}.
\newblock \bibinfo{journal}{\emph{ACM TOG}} (\bibinfo{year}{2021}).
\newblock


\bibitem[Anokhin et~al\mbox{.}(2021)]%
        {anokhin2020image}
\bibfield{author}{\bibinfo{person}{Ivan Anokhin}, \bibinfo{person}{Kirill Demochkin}, \bibinfo{person}{Taras Khakhulin}, \bibinfo{person}{Gleb Sterkin}, \bibinfo{person}{Victor Lempitsky}, {and} \bibinfo{person}{Denis Korzhenkov}.} \bibinfo{year}{2021}\natexlab{}.
\newblock \showarticletitle{Image Generators with Conditionally-Independent Pixel Synthesis}. In \bibinfo{booktitle}{\emph{CVPR}}.
\newblock


\bibitem[Arandjelovi{\'c} and Zisserman(2012)]%
        {arandjelovic2012three}
\bibfield{author}{\bibinfo{person}{Relja Arandjelovi{\'c}} {and} \bibinfo{person}{Andrew Zisserman}.} \bibinfo{year}{2012}\natexlab{}.
\newblock \showarticletitle{Three things everyone should know to improve object retrieval}. In \bibinfo{booktitle}{\emph{CVPR}}.
\newblock


\bibitem[Au(2019)]%
        {au2019vesselgan}
\bibfield{author}{\bibinfo{person}{Derek~Philip Au}.} \bibinfo{year}{2019}\natexlab{}.
\newblock \bibinfo{title}{This vessel does not exist.}
\newblock \bibinfo{howpublished}{https://thisvesseldoesnotexist.com/}.
\newblock


\bibitem[Avrahami et~al\mbox{.}(2022)]%
        {avrahami2022blended}
\bibfield{author}{\bibinfo{person}{Omri Avrahami}, \bibinfo{person}{Dani Lischinski}, {and} \bibinfo{person}{Ohad Fried}.} \bibinfo{year}{2022}\natexlab{}.
\newblock \showarticletitle{Blended diffusion for text-driven editing of natural images}. In \bibinfo{booktitle}{\emph{CVPR}}.
\newblock


\bibitem[Babenko et~al\mbox{.}(2014)]%
        {babenko2014neural}
\bibfield{author}{\bibinfo{person}{Artem Babenko}, \bibinfo{person}{Anton Slesarev}, \bibinfo{person}{Alexandr Chigorin}, {and} \bibinfo{person}{Victor Lempitsky}.} \bibinfo{year}{2014}\natexlab{}.
\newblock \showarticletitle{Neural codes for image retrieval}. In \bibinfo{booktitle}{\emph{ECCV}}.
\newblock


\bibitem[Baeza-Yates et~al\mbox{.}(1999)]%
        {baeza1999modern}
\bibfield{author}{\bibinfo{person}{Ricardo Baeza-Yates}, \bibinfo{person}{Berthier Ribeiro-Neto}, {et~al\mbox{.}}} \bibinfo{year}{1999}\natexlab{}.
\newblock \bibinfo{booktitle}{\emph{Modern information retrieval}}. Vol.~\bibinfo{volume}{463}.
\newblock \bibinfo{publisher}{ACM press New York}.
\newblock


\bibitem[Bau et~al\mbox{.}(2020)]%
        {bau2020rewriting}
\bibfield{author}{\bibinfo{person}{David Bau}, \bibinfo{person}{Steven Liu}, \bibinfo{person}{Tongzhou Wang}, \bibinfo{person}{Jun-Yan Zhu}, {and} \bibinfo{person}{Antonio Torralba}.} \bibinfo{year}{2020}\natexlab{}.
\newblock \showarticletitle{Rewriting a deep generative model}. In \bibinfo{booktitle}{\emph{ECCV}}.
\newblock


\bibitem[Bermano et~al\mbox{.}(2022)]%
        {bermano2022state}
\bibfield{author}{\bibinfo{person}{Amit~H Bermano}, \bibinfo{person}{Rinon Gal}, \bibinfo{person}{Yuval Alaluf}, \bibinfo{person}{Ron Mokady}, \bibinfo{person}{Yotam Nitzan}, \bibinfo{person}{Omer Tov}, \bibinfo{person}{Or Patashnik}, {and} \bibinfo{person}{Daniel Cohen-Or}.} \bibinfo{year}{2022}\natexlab{}.
\newblock \showarticletitle{State-of-the-Art in the Architecture, Methods and Applications of StyleGAN}.
\newblock \bibinfo{journal}{\emph{arXiv preprint arXiv:2202.14020}} (\bibinfo{year}{2022}).
\newblock


\bibitem[Blattmann et~al\mbox{.}(2022)]%
        {blattmann2022retrieval}
\bibfield{author}{\bibinfo{person}{Andreas Blattmann}, \bibinfo{person}{Robin Rombach}, \bibinfo{person}{Kaan Oktay}, \bibinfo{person}{Jonas M{\"u}ller}, {and} \bibinfo{person}{Bj{\"o}rn Ommer}.} \bibinfo{year}{2022}\natexlab{}.
\newblock \showarticletitle{Retrieval-augmented diffusion models}.
\newblock \bibinfo{journal}{\emph{Advances in Neural Information Processing Systems}}  \bibinfo{volume}{35} (\bibinfo{year}{2022}), \bibinfo{pages}{15309--15324}.
\newblock


\bibitem[Brock et~al\mbox{.}(2019)]%
        {brock2019large}
\bibfield{author}{\bibinfo{person}{Andrew Brock}, \bibinfo{person}{Jeff Donahue}, {and} \bibinfo{person}{Karen Simonyan}.} \bibinfo{year}{2019}\natexlab{}.
\newblock \showarticletitle{Large scale gan training for high fidelity natural image synthesis}. In \bibinfo{booktitle}{\emph{ICLR}}.
\newblock


\bibitem[Brooks et~al\mbox{.}(2023)]%
        {brooks2022instructpix2pix}
\bibfield{author}{\bibinfo{person}{Tim Brooks}, \bibinfo{person}{Aleksander Holynski}, {and} \bibinfo{person}{Alexei~A Efros}.} \bibinfo{year}{2023}\natexlab{}.
\newblock \showarticletitle{Instructpix2pix: Learning to follow image editing instructions}. In \bibinfo{booktitle}{\emph{Proceedings of the IEEE/CVF Conference on Computer Vision and Pattern Recognition}}. \bibinfo{pages}{18392--18402}.
\newblock


\bibitem[Casanova et~al\mbox{.}(2021)]%
        {casanova2021instance}
\bibfield{author}{\bibinfo{person}{Arantxa Casanova}, \bibinfo{person}{Marlene Careil}, \bibinfo{person}{Jakob Verbeek}, \bibinfo{person}{Michal Drozdzal}, {and} \bibinfo{person}{Adriana Romero~Soriano}.} \bibinfo{year}{2021}\natexlab{}.
\newblock \showarticletitle{Instance-conditioned gan}.
\newblock \bibinfo{journal}{\emph{Advances in Neural Information Processing Systems}}  \bibinfo{volume}{34} (\bibinfo{year}{2021}), \bibinfo{pages}{27517--27529}.
\newblock


\bibitem[Chan et~al\mbox{.}(2022)]%
        {chan2022learning}
\bibfield{author}{\bibinfo{person}{Caroline Chan}, \bibinfo{person}{Fredo Durand}, {and} \bibinfo{person}{Phillip Isola}.} \bibinfo{year}{2022}\natexlab{}.
\newblock \showarticletitle{Learning to generate line drawings that convey geometry and semantics}. In \bibinfo{booktitle}{\emph{CVPR}}.
\newblock


\bibitem[Chen et~al\mbox{.}(2020)]%
        {chen2020deepfacedrawing}
\bibfield{author}{\bibinfo{person}{Shu-Yu Chen}, \bibinfo{person}{Wanchao Su}, \bibinfo{person}{Lin Gao}, \bibinfo{person}{Shihong Xia}, {and} \bibinfo{person}{Hongbo Fu}.} \bibinfo{year}{2020}\natexlab{}.
\newblock \showarticletitle{DeepFaceDrawing: Deep generation of face images from sketches}.
\newblock \bibinfo{journal}{\emph{ACM Transactions on Graphics (TOG)}} \bibinfo{volume}{39}, \bibinfo{number}{4} (\bibinfo{year}{2020}), \bibinfo{pages}{72--1}.
\newblock


\bibitem[Chen et~al\mbox{.}(2023)]%
        {chen2022re}
\bibfield{author}{\bibinfo{person}{Wenhu Chen}, \bibinfo{person}{Hexiang Hu}, \bibinfo{person}{Chitwan Saharia}, {and} \bibinfo{person}{William~W Cohen}.} \bibinfo{year}{2023}\natexlab{}.
\newblock \showarticletitle{Re-imagen: Retrieval-augmented text-to-image generator}. In \bibinfo{booktitle}{\emph{ICLR}}.
\newblock


\bibitem[Choi et~al\mbox{.}(2020)]%
        {choi2020starganv2}
\bibfield{author}{\bibinfo{person}{Yunjey Choi}, \bibinfo{person}{Youngjung Uh}, \bibinfo{person}{Jaejun Yoo}, {and} \bibinfo{person}{Jung-Woo Ha}.} \bibinfo{year}{2020}\natexlab{}.
\newblock \showarticletitle{StarGAN v2: Diverse Image Synthesis for Multiple Domains}. In \bibinfo{booktitle}{\emph{CVPR}}.
\newblock


\bibitem[Dalal and Triggs(2005)]%
        {dalal2005histograms}
\bibfield{author}{\bibinfo{person}{Navneet Dalal} {and} \bibinfo{person}{Bill Triggs}.} \bibinfo{year}{2005}\natexlab{}.
\newblock \showarticletitle{Histograms of oriented gradients for human detection}. In \bibinfo{booktitle}{\emph{CVPR}}.
\newblock


\bibitem[Datta et~al\mbox{.}(2008)]%
        {datta2008image}
\bibfield{author}{\bibinfo{person}{Ritendra Datta}, \bibinfo{person}{Dhiraj Joshi}, \bibinfo{person}{Jia Li}, {and} \bibinfo{person}{James~Z Wang}.} \bibinfo{year}{2008}\natexlab{}.
\newblock \showarticletitle{Image retrieval: Ideas, influences, and trends of the new age}.
\newblock \bibinfo{journal}{\emph{ACM Computing Surveys (Csur)}} (\bibinfo{year}{2008}).
\newblock


\bibitem[Deng et~al\mbox{.}(2009)]%
        {deng2009imagenet}
\bibfield{author}{\bibinfo{person}{J. Deng}, \bibinfo{person}{W. Dong}, \bibinfo{person}{R. Socher}, \bibinfo{person}{L.-J. Li}, \bibinfo{person}{K. Li}, {and} \bibinfo{person}{L. Fei-Fei}.} \bibinfo{year}{2009}\natexlab{}.
\newblock \showarticletitle{{ImageNet: A Large-Scale Hierarchical Image Database}}. In \bibinfo{booktitle}{\emph{CVPR}}.
\newblock


\bibitem[Dhariwal and Nichol(2021)]%
        {dhariwal2021diffusion}
\bibfield{author}{\bibinfo{person}{Prafulla Dhariwal} {and} \bibinfo{person}{Alexander Nichol}.} \bibinfo{year}{2021}\natexlab{}.
\newblock \showarticletitle{Diffusion models beat gans on image synthesis}. In \bibinfo{booktitle}{\emph{NeurIPS}}.
\newblock


\bibitem[Eitz et~al\mbox{.}(2010)]%
        {eitz2010sketch}
\bibfield{author}{\bibinfo{person}{Mathias Eitz}, \bibinfo{person}{Kristian Hildebrand}, \bibinfo{person}{Tamy Boubekeur}, {and} \bibinfo{person}{Marc Alexa}.} \bibinfo{year}{2010}\natexlab{}.
\newblock \showarticletitle{Sketch-based image retrieval: Benchmark and bag-of-features descriptors}.
\newblock \bibinfo{journal}{\emph{IEEE transactions on visualization and computer graphics}} \bibinfo{volume}{17}, \bibinfo{number}{11} (\bibinfo{year}{2010}), \bibinfo{pages}{1624--1636}.
\newblock


\bibitem[Elgammal(2019)]%
        {elgammal2019ai}
\bibfield{author}{\bibinfo{person}{Ahmed Elgammal}.} \bibinfo{year}{2019}\natexlab{}.
\newblock \showarticletitle{AI is blurring the definition of artist: Advanced algorithms are using machine learning to create art autonomously}.
\newblock \bibinfo{journal}{\emph{American Scientist}} \bibinfo{volume}{107}, \bibinfo{number}{1} (\bibinfo{year}{2019}), \bibinfo{pages}{18--22}.
\newblock


\bibitem[Esser et~al\mbox{.}(2021)]%
        {esser2021taming}
\bibfield{author}{\bibinfo{person}{Patrick Esser}, \bibinfo{person}{Robin Rombach}, {and} \bibinfo{person}{Bjorn Ommer}.} \bibinfo{year}{2021}\natexlab{}.
\newblock \showarticletitle{Taming transformers for high-resolution image synthesis}. In \bibinfo{booktitle}{\emph{CVPR}}.
\newblock


\bibitem[Faghri et~al\mbox{.}(2017)]%
        {faghri2017vse}
\bibfield{author}{\bibinfo{person}{Fartash Faghri}, \bibinfo{person}{David~J Fleet}, \bibinfo{person}{Jamie~Ryan Kiros}, {and} \bibinfo{person}{Sanja Fidler}.} \bibinfo{year}{2017}\natexlab{}.
\newblock \showarticletitle{Vse++: Improving visual-semantic embeddings with hard negatives}. In \bibinfo{booktitle}{\emph{BMVC}}.
\newblock


\bibitem[Frome et~al\mbox{.}(2013)]%
        {frome2013devise}
\bibfield{author}{\bibinfo{person}{Andrea Frome}, \bibinfo{person}{Greg~S Corrado}, \bibinfo{person}{Jon Shlens}, \bibinfo{person}{Samy Bengio}, \bibinfo{person}{Jeff Dean}, \bibinfo{person}{Marc'Aurelio Ranzato}, {and} \bibinfo{person}{Tomas Mikolov}.} \bibinfo{year}{2013}\natexlab{}.
\newblock \showarticletitle{Devise: A deep visual-semantic embedding model}. In \bibinfo{booktitle}{\emph{NeurIPS}}.
\newblock


\bibitem[Gal et~al\mbox{.}(2022a)]%
        {gal2022image}
\bibfield{author}{\bibinfo{person}{Rinon Gal}, \bibinfo{person}{Yuval Alaluf}, \bibinfo{person}{Yuval Atzmon}, \bibinfo{person}{Or Patashnik}, \bibinfo{person}{Amit~H Bermano}, \bibinfo{person}{Gal Chechik}, {and} \bibinfo{person}{Daniel Cohen-Or}.} \bibinfo{year}{2022}\natexlab{a}.
\newblock \showarticletitle{An image is worth one word: Personalizing text-to-image generation using textual inversion}.
\newblock \bibinfo{journal}{\emph{arXiv preprint arXiv:2208.01618}} (\bibinfo{year}{2022}).
\newblock


\bibitem[Gal et~al\mbox{.}(2023)]%
        {gal2023designing}
\bibfield{author}{\bibinfo{person}{Rinon Gal}, \bibinfo{person}{Moab Arar}, \bibinfo{person}{Yuval Atzmon}, \bibinfo{person}{Amit~H Bermano}, \bibinfo{person}{Gal Chechik}, {and} \bibinfo{person}{Daniel Cohen-Or}.} \bibinfo{year}{2023}\natexlab{}.
\newblock \showarticletitle{Encoder-based domain tuning for fast personalization of text-to-image models}.
\newblock \bibinfo{journal}{\emph{ACM Transactions on Graphics (TOG)}} \bibinfo{volume}{42}, \bibinfo{number}{4} (\bibinfo{year}{2023}), \bibinfo{pages}{1--13}.
\newblock


\bibitem[Gal et~al\mbox{.}(2022b)]%
        {gal2021stylegan}
\bibfield{author}{\bibinfo{person}{Rinon Gal}, \bibinfo{person}{Or Patashnik}, \bibinfo{person}{Haggai Maron}, \bibinfo{person}{Gal Chechik}, {and} \bibinfo{person}{Daniel Cohen-Or}.} \bibinfo{year}{2022}\natexlab{b}.
\newblock \showarticletitle{StyleGAN-NADA: CLIP-Guided Domain Adaptation of Image Generators}.
\newblock \bibinfo{journal}{\emph{ACM TOG}} (\bibinfo{year}{2022}).
\newblock


\bibitem[Gong et~al\mbox{.}(2012)]%
        {gong2012iterative}
\bibfield{author}{\bibinfo{person}{Yunchao Gong}, \bibinfo{person}{Svetlana Lazebnik}, \bibinfo{person}{Albert Gordo}, {and} \bibinfo{person}{Florent Perronnin}.} \bibinfo{year}{2012}\natexlab{}.
\newblock \showarticletitle{Iterative quantization: A procrustean approach to learning binary codes for large-scale image retrieval}.
\newblock \bibinfo{journal}{\emph{IEEE transactions on pattern analysis and machine intelligence}} \bibinfo{volume}{35}, \bibinfo{number}{12} (\bibinfo{year}{2012}), \bibinfo{pages}{2916--2929}.
\newblock


\bibitem[Goodfellow et~al\mbox{.}(2014)]%
        {goodfellow2014generative}
\bibfield{author}{\bibinfo{person}{Ian Goodfellow}, \bibinfo{person}{Jean Pouget-Abadie}, \bibinfo{person}{Mehdi Mirza}, \bibinfo{person}{Bing Xu}, \bibinfo{person}{David Warde-Farley}, \bibinfo{person}{Sherjil Ozair}, \bibinfo{person}{Aaron Courville}, {and} \bibinfo{person}{Yoshua Bengio}.} \bibinfo{year}{2014}\natexlab{}.
\newblock \showarticletitle{Generative adversarial nets}. In \bibinfo{booktitle}{\emph{NeurIPS}}.
\newblock


\bibitem[Grigoryev et~al\mbox{.}(2022)]%
        {grigoryev2022and}
\bibfield{author}{\bibinfo{person}{Timofey Grigoryev}, \bibinfo{person}{Andrey Voynov}, {and} \bibinfo{person}{Artem Babenko}.} \bibinfo{year}{2022}\natexlab{}.
\newblock \showarticletitle{When, Why, and Which Pretrained GANs Are Useful?}. In \bibinfo{booktitle}{\emph{ICLR}}.
\newblock


\bibitem[Gudivada and Raghavan(1995)]%
        {gudivada1995content}
\bibfield{author}{\bibinfo{person}{Venkat~N Gudivada} {and} \bibinfo{person}{Vijay~V Raghavan}.} \bibinfo{year}{1995}\natexlab{}.
\newblock \showarticletitle{Content based image retrieval systems}.
\newblock \bibinfo{journal}{\emph{Computer}} \bibinfo{volume}{28}, \bibinfo{number}{9} (\bibinfo{year}{1995}), \bibinfo{pages}{18--22}.
\newblock


\bibitem[Ha and Schmidhuber(2018)]%
        {ha2018world}
\bibfield{author}{\bibinfo{person}{David Ha} {and} \bibinfo{person}{J{\"u}rgen Schmidhuber}.} \bibinfo{year}{2018}\natexlab{}.
\newblock \showarticletitle{World models}.
\newblock \bibinfo{journal}{\emph{arXiv preprint arXiv:1803.10122}} (\bibinfo{year}{2018}).
\newblock


\bibitem[Han et~al\mbox{.}(2023)]%
        {han2023svdiff}
\bibfield{author}{\bibinfo{person}{Ligong Han}, \bibinfo{person}{Yinxiao Li}, \bibinfo{person}{Han Zhang}, \bibinfo{person}{Peyman Milanfar}, \bibinfo{person}{Dimitris Metaxas}, {and} \bibinfo{person}{Feng Yang}.} \bibinfo{year}{2023}\natexlab{}.
\newblock \showarticletitle{Svdiff: Compact parameter space for diffusion fine-tuning}.
\newblock \bibinfo{journal}{\emph{arXiv preprint arXiv:2303.11305}} (\bibinfo{year}{2023}).
\newblock


\bibitem[Hertz et~al\mbox{.}(2022)]%
        {hertz2022prompt}
\bibfield{author}{\bibinfo{person}{Amir Hertz}, \bibinfo{person}{Ron Mokady}, \bibinfo{person}{Jay Tenenbaum}, \bibinfo{person}{Kfir Aberman}, \bibinfo{person}{Yael Pritch}, {and} \bibinfo{person}{Daniel Cohen-Or}.} \bibinfo{year}{2022}\natexlab{}.
\newblock \showarticletitle{Prompt-to-prompt image editing with cross attention control}.
\newblock \bibinfo{journal}{\emph{arXiv preprint arXiv:2208.01626}} (\bibinfo{year}{2022}).
\newblock


\bibitem[Hertzmann(2020)]%
        {hertzmann2020computers}
\bibfield{author}{\bibinfo{person}{Aaron Hertzmann}.} \bibinfo{year}{2020}\natexlab{}.
\newblock \showarticletitle{Computers do not make art, people do}.
\newblock \bibinfo{journal}{\emph{Commun. ACM}} \bibinfo{volume}{63}, \bibinfo{number}{5} (\bibinfo{year}{2020}), \bibinfo{pages}{45--48}.
\newblock


\bibitem[Heusel et~al\mbox{.}(2017)]%
        {heusel2017gans}
\bibfield{author}{\bibinfo{person}{Martin Heusel}, \bibinfo{person}{Hubert Ramsauer}, \bibinfo{person}{Thomas Unterthiner}, \bibinfo{person}{Bernhard Nessler}, {and} \bibinfo{person}{Sepp Hochreiter}.} \bibinfo{year}{2017}\natexlab{}.
\newblock \showarticletitle{{GANs} trained by a two time-scale update rule converge to a local {Nash} equilibrium}. In \bibinfo{booktitle}{\emph{NeurIPS}}.
\newblock


\bibitem[Hinton(2002)]%
        {hinton2002training}
\bibfield{author}{\bibinfo{person}{Geoffrey~E Hinton}.} \bibinfo{year}{2002}\natexlab{}.
\newblock \showarticletitle{Training products of experts by minimizing contrastive divergence}.
\newblock \bibinfo{journal}{\emph{Neural computation}} \bibinfo{volume}{14}, \bibinfo{number}{8} (\bibinfo{year}{2002}), \bibinfo{pages}{1771--1800}.
\newblock


\bibitem[Ho et~al\mbox{.}(2020)]%
        {ho2020denoising}
\bibfield{author}{\bibinfo{person}{Jonathan Ho}, \bibinfo{person}{Ajay Jain}, {and} \bibinfo{person}{Pieter Abbeel}.} \bibinfo{year}{2020}\natexlab{}.
\newblock \showarticletitle{Denoising Diffusion Probabilistic Models}. In \bibinfo{booktitle}{\emph{NeurIPS}}.
\newblock


\bibitem[Hu et~al\mbox{.}(2021)]%
        {hu2021lora}
\bibfield{author}{\bibinfo{person}{Edward~J Hu}, \bibinfo{person}{Yelong Shen}, \bibinfo{person}{Phillip Wallis}, \bibinfo{person}{Zeyuan Allen-Zhu}, \bibinfo{person}{Yuanzhi Li}, \bibinfo{person}{Shean Wang}, \bibinfo{person}{Lu Wang}, {and} \bibinfo{person}{Weizhu Chen}.} \bibinfo{year}{2021}\natexlab{}.
\newblock \showarticletitle{Lora: Low-rank adaptation of large language models}.
\newblock \bibinfo{journal}{\emph{arXiv preprint arXiv:2106.09685}} (\bibinfo{year}{2021}).
\newblock


\bibitem[Hu et~al\mbox{.}(2011)]%
        {hu2011survey}
\bibfield{author}{\bibinfo{person}{Weiming Hu}, \bibinfo{person}{Nianhua Xie}, \bibinfo{person}{Li Li}, \bibinfo{person}{Xianglin Zeng}, {and} \bibinfo{person}{Stephen Maybank}.} \bibinfo{year}{2011}\natexlab{}.
\newblock \showarticletitle{A survey on visual content-based video indexing and retrieval}.
\newblock \bibinfo{journal}{\emph{IEEE Transactions on Systems, Man, and Cybernetics, Part C (Applications and Reviews)}} \bibinfo{volume}{41}, \bibinfo{number}{6} (\bibinfo{year}{2011}), \bibinfo{pages}{797--819}.
\newblock


\bibitem[Huang et~al\mbox{.}(2022)]%
        {huang2021multimodal}
\bibfield{author}{\bibinfo{person}{Xun Huang}, \bibinfo{person}{Arun Mallya}, \bibinfo{person}{Ting-Chun Wang}, {and} \bibinfo{person}{Ming-Yu Liu}.} \bibinfo{year}{2022}\natexlab{}.
\newblock \showarticletitle{Multimodal Conditional Image Synthesis with Product-of-Experts GANs}. In \bibinfo{booktitle}{\emph{ECCV}}.
\newblock


\bibitem[Jang et~al\mbox{.}(2021)]%
        {jang2021stylecari}
\bibfield{author}{\bibinfo{person}{Wonjong Jang}, \bibinfo{person}{Gwangjin Ju}, \bibinfo{person}{Yucheol Jung}, \bibinfo{person}{Jiaolong Yang}, \bibinfo{person}{Xin Tong}, {and} \bibinfo{person}{Seungyong Lee}.} \bibinfo{year}{2021}\natexlab{}.
\newblock \showarticletitle{StyleCariGAN: Caricature Generation via StyleGAN Feature Map Modulation}.
\newblock \bibinfo{journal}{\emph{ACM TOG}} (\bibinfo{year}{2021}).
\newblock


\bibitem[J{\'e}gou et~al\mbox{.}(2010)]%
        {jegou2010aggregating}
\bibfield{author}{\bibinfo{person}{Herv{\'e} J{\'e}gou}, \bibinfo{person}{Matthijs Douze}, \bibinfo{person}{Cordelia Schmid}, {and} \bibinfo{person}{Patrick P{\'e}rez}.} \bibinfo{year}{2010}\natexlab{}.
\newblock \showarticletitle{Aggregating local descriptors into a compact image representation}. In \bibinfo{booktitle}{\emph{2010 IEEE computer society conference on computer vision and pattern recognition}}. IEEE, \bibinfo{pages}{3304--3311}.
\newblock


\bibitem[Jia et~al\mbox{.}(2021)]%
        {jia2021scaling}
\bibfield{author}{\bibinfo{person}{Chao Jia}, \bibinfo{person}{Yinfei Yang}, \bibinfo{person}{Ye Xia}, \bibinfo{person}{Yi-Ting Chen}, \bibinfo{person}{Zarana Parekh}, \bibinfo{person}{Hieu Pham}, \bibinfo{person}{Quoc Le}, \bibinfo{person}{Yun-Hsuan Sung}, \bibinfo{person}{Zhen Li}, {and} \bibinfo{person}{Tom Duerig}.} \bibinfo{year}{2021}\natexlab{}.
\newblock \showarticletitle{Scaling up visual and vision-language representation learning with noisy text supervision}. In \bibinfo{booktitle}{\emph{ICML}}.
\newblock


\bibitem[Karpathy et~al\mbox{.}(2014)]%
        {karpathy2014deep}
\bibfield{author}{\bibinfo{person}{Andrej Karpathy}, \bibinfo{person}{Armand Joulin}, {and} \bibinfo{person}{Li~F Fei-Fei}.} \bibinfo{year}{2014}\natexlab{}.
\newblock \showarticletitle{Deep fragment embeddings for bidirectional image sentence mapping}. In \bibinfo{booktitle}{\emph{NeurIPS}}.
\newblock


\bibitem[Karras et~al\mbox{.}(2018)]%
        {karras2018progressive}
\bibfield{author}{\bibinfo{person}{Tero Karras}, \bibinfo{person}{Timo Aila}, \bibinfo{person}{Samuli Laine}, {and} \bibinfo{person}{Jaakko Lehtinen}.} \bibinfo{year}{2018}\natexlab{}.
\newblock \showarticletitle{Progressive growing of gans for improved quality, stability, and variation}. In \bibinfo{booktitle}{\emph{ICLR}}.
\newblock


\bibitem[Karras et~al\mbox{.}(2020a)]%
        {karras2020ADA}
\bibfield{author}{\bibinfo{person}{Tero Karras}, \bibinfo{person}{Miika Aittala}, \bibinfo{person}{Janne Hellsten}, \bibinfo{person}{Samuli Laine}, \bibinfo{person}{Jaakko Lehtinen}, {and} \bibinfo{person}{Timo Aila}.} \bibinfo{year}{2020}\natexlab{a}.
\newblock \showarticletitle{Training Generative Adversarial Networks with Limited Data}. In \bibinfo{booktitle}{\emph{NeurIPS}}.
\newblock


\bibitem[Karras et~al\mbox{.}(2021)]%
        {Karras2021alias}
\bibfield{author}{\bibinfo{person}{Tero Karras}, \bibinfo{person}{Miika Aittala}, \bibinfo{person}{Samuli Laine}, \bibinfo{person}{Erik H\"ark\"onen}, \bibinfo{person}{Janne Hellsten}, \bibinfo{person}{Jaakko Lehtinen}, {and} \bibinfo{person}{Timo Aila}.} \bibinfo{year}{2021}\natexlab{}.
\newblock \showarticletitle{Alias-Free Generative Adversarial Networks}. In \bibinfo{booktitle}{\emph{NeurIPS}}.
\newblock


\bibitem[Karras et~al\mbox{.}(2019)]%
        {karras2019style}
\bibfield{author}{\bibinfo{person}{Tero Karras}, \bibinfo{person}{Samuli Laine}, {and} \bibinfo{person}{Timo Aila}.} \bibinfo{year}{2019}\natexlab{}.
\newblock \showarticletitle{A style-based generator architecture for generative adversarial networks}. In \bibinfo{booktitle}{\emph{CVPR}}.
\newblock


\bibitem[Karras et~al\mbox{.}(2020b)]%
        {karras2020analyzing}
\bibfield{author}{\bibinfo{person}{Tero Karras}, \bibinfo{person}{Samuli Laine}, \bibinfo{person}{Miika Aittala}, \bibinfo{person}{Janne Hellsten}, \bibinfo{person}{Jaakko Lehtinen}, {and} \bibinfo{person}{Timo Aila}.} \bibinfo{year}{2020}\natexlab{b}.
\newblock \showarticletitle{Analyzing and improving the image quality of stylegan}. In \bibinfo{booktitle}{\emph{CVPR}}.
\newblock


\bibitem[Kawar et~al\mbox{.}(2023)]%
        {kawar2022imagic}
\bibfield{author}{\bibinfo{person}{Bahjat Kawar}, \bibinfo{person}{Shiran Zada}, \bibinfo{person}{Oran Lang}, \bibinfo{person}{Omer Tov}, \bibinfo{person}{Huiwen Chang}, \bibinfo{person}{Tali Dekel}, \bibinfo{person}{Inbar Mosseri}, {and} \bibinfo{person}{Michal Irani}.} \bibinfo{year}{2023}\natexlab{}.
\newblock \showarticletitle{Imagic: Text-based real image editing with diffusion models}.
\newblock  (\bibinfo{year}{2023}), \bibinfo{pages}{6007--6017}.
\newblock


\bibitem[Kingma and Welling(2014)]%
        {kingma2013auto}
\bibfield{author}{\bibinfo{person}{Diederik~P Kingma} {and} \bibinfo{person}{Max Welling}.} \bibinfo{year}{2014}\natexlab{}.
\newblock \showarticletitle{Auto-encoding variational bayes}. In \bibinfo{booktitle}{\emph{ICLR}}.
\newblock


\bibitem[Krizhevsky and Hinton(2011)]%
        {krizhevsky2011using}
\bibfield{author}{\bibinfo{person}{Alex Krizhevsky} {and} \bibinfo{person}{Geoffrey~E Hinton}.} \bibinfo{year}{2011}\natexlab{}.
\newblock \showarticletitle{Using very deep autoencoders for content-based image retrieval.}. In \bibinfo{booktitle}{\emph{ESANN}}, Vol.~\bibinfo{volume}{1}. Citeseer, \bibinfo{pages}{2}.
\newblock


\bibitem[Kumari et~al\mbox{.}(2023)]%
        {kumari2022multi}
\bibfield{author}{\bibinfo{person}{Nupur Kumari}, \bibinfo{person}{Bingliang Zhang}, \bibinfo{person}{Richard Zhang}, \bibinfo{person}{Eli Shechtman}, {and} \bibinfo{person}{Jun-Yan Zhu}.} \bibinfo{year}{2023}\natexlab{}.
\newblock \showarticletitle{Multi-concept customization of text-to-image diffusion}.
\newblock  (\bibinfo{year}{2023}), \bibinfo{pages}{1931--1941}.
\newblock


\bibitem[Kumari et~al\mbox{.}(2022)]%
        {kumari2021ensembling}
\bibfield{author}{\bibinfo{person}{Nupur Kumari}, \bibinfo{person}{Richard Zhang}, \bibinfo{person}{Eli Shechtman}, {and} \bibinfo{person}{Jun-Yan Zhu}.} \bibinfo{year}{2022}\natexlab{}.
\newblock \showarticletitle{Ensembling Off-the-shelf Models for GAN Training}. In \bibinfo{booktitle}{\emph{CVPR}}.
\newblock


\bibitem[Lewis et~al\mbox{.}(2021)]%
        {lewis2021tryongan}
\bibfield{author}{\bibinfo{person}{Kathleen~M Lewis}, \bibinfo{person}{Srivatsan Varadharajan}, {and} \bibinfo{person}{Ira Kemelmacher-Shlizerman}.} \bibinfo{year}{2021}\natexlab{}.
\newblock \showarticletitle{TryOnGAN: Body-Aware Try-On via Layered Interpolation}.
\newblock \bibinfo{journal}{\emph{ACM TOG}} (\bibinfo{year}{2021}).
\newblock


\bibitem[Li et~al\mbox{.}(2022)]%
        {li2022blip}
\bibfield{author}{\bibinfo{person}{Junnan Li}, \bibinfo{person}{Dongxu Li}, \bibinfo{person}{Caiming Xiong}, {and} \bibinfo{person}{Steven Hoi}.} \bibinfo{year}{2022}\natexlab{}.
\newblock \bibinfo{title}{BLIP: Bootstrapping Language-Image Pre-training for Unified Vision-Language Understanding and Generation}.
\newblock
\newblock
\showeprint[arxiv]{2201.12086}~[cs.CV]


\bibitem[Li et~al\mbox{.}(2020)]%
        {li2020few}
\bibfield{author}{\bibinfo{person}{Yijun Li}, \bibinfo{person}{Richard Zhang}, \bibinfo{person}{Jingwan Lu}, {and} \bibinfo{person}{Eli Shechtman}.} \bibinfo{year}{2020}\natexlab{}.
\newblock \showarticletitle{Few-shot image generation with elastic weight consolidation}. In \bibinfo{booktitle}{\emph{NeurIPS}}.
\newblock


\bibitem[Lin et~al\mbox{.}(2013)]%
        {lin20133d}
\bibfield{author}{\bibinfo{person}{Yen-Liang Lin}, \bibinfo{person}{Cheng-Yu Huang}, \bibinfo{person}{Hao-Jeng Wang}, {and} \bibinfo{person}{Winston Hsu}.} \bibinfo{year}{2013}\natexlab{}.
\newblock \showarticletitle{3D sub-query expansion for improving sketch-based multi-view image retrieval}. In \bibinfo{booktitle}{\emph{ICCV}}.
\newblock


\bibitem[Liu et~al\mbox{.}(2021)]%
        {liu2020towards}
\bibfield{author}{\bibinfo{person}{Bingchen Liu}, \bibinfo{person}{Yizhe Zhu}, \bibinfo{person}{Kunpeng Song}, {and} \bibinfo{person}{Ahmed Elgammal}.} \bibinfo{year}{2021}\natexlab{}.
\newblock \showarticletitle{Towards faster and stabilized gan training for high-fidelity few-shot image synthesis}. In \bibinfo{booktitle}{\emph{ICLR}}.
\newblock


\bibitem[Liu et~al\mbox{.}(2017)]%
        {liu2017deep}
\bibfield{author}{\bibinfo{person}{Li Liu}, \bibinfo{person}{Fumin Shen}, \bibinfo{person}{Yuming Shen}, \bibinfo{person}{Xianglong Liu}, {and} \bibinfo{person}{Ling Shao}.} \bibinfo{year}{2017}\natexlab{}.
\newblock \showarticletitle{Deep sketch hashing: Fast free-hand sketch-based image retrieval}. In \bibinfo{booktitle}{\emph{CVPR}}.
\newblock


\bibitem[Liu et~al\mbox{.}(2022)]%
        {liu2022convnet}
\bibfield{author}{\bibinfo{person}{Zhuang Liu}, \bibinfo{person}{Hanzi Mao}, \bibinfo{person}{Chao-Yuan Wu}, \bibinfo{person}{Christoph Feichtenhofer}, \bibinfo{person}{Trevor Darrell}, {and} \bibinfo{person}{Saining Xie}.} \bibinfo{year}{2022}\natexlab{}.
\newblock \showarticletitle{A convnet for the 2020s}. In \bibinfo{booktitle}{\emph{Proceedings of the IEEE/CVF conference on computer vision and pattern recognition}}. \bibinfo{pages}{11976--11986}.
\newblock


\bibitem[Lowe(2004)]%
        {lowe2004distinctive}
\bibfield{author}{\bibinfo{person}{David~G Lowe}.} \bibinfo{year}{2004}\natexlab{}.
\newblock \showarticletitle{Distinctive image features from scale-invariant keypoints}.
\newblock \bibinfo{journal}{\emph{IJCV}} \bibinfo{volume}{60}, \bibinfo{number}{2} (\bibinfo{year}{2004}), \bibinfo{pages}{91--110}.
\newblock


\bibitem[lucid layers(2022)]%
        {lucidlayersstylegan3}
\bibfield{author}{\bibinfo{person}{lucid layers}.} \bibinfo{year}{2022}\natexlab{}.
\newblock \bibinfo{title}{Datasets and pretrained Models for StyleGAN3}.
\newblock \bibinfo{howpublished}{https://github.com/edstoica/lucid\_stylegan3\_datasets\_models/}.
\newblock


\bibitem[Ma et~al\mbox{.}(2023)]%
        {ma2023unified}
\bibfield{author}{\bibinfo{person}{Yiyang Ma}, \bibinfo{person}{Huan Yang}, \bibinfo{person}{Wenjing Wang}, \bibinfo{person}{Jianlong Fu}, {and} \bibinfo{person}{Jiaying Liu}.} \bibinfo{year}{2023}\natexlab{}.
\newblock \showarticletitle{Unified multi-modal latent diffusion for joint subject and text conditional image generation}.
\newblock \bibinfo{journal}{\emph{arXiv preprint arXiv:2303.09319}} (\bibinfo{year}{2023}).
\newblock


\bibitem[Manning et~al\mbox{.}(2010)]%
        {manning2010introduction}
\bibfield{author}{\bibinfo{person}{Christopher Manning}, \bibinfo{person}{Prabhakar Raghavan}, {and} \bibinfo{person}{Hinrich Sch{\"u}tze}.} \bibinfo{year}{2010}\natexlab{}.
\newblock \showarticletitle{Introduction to information retrieval}.
\newblock \bibinfo{journal}{\emph{Natural Language Engineering}} \bibinfo{volume}{16}, \bibinfo{number}{1} (\bibinfo{year}{2010}), \bibinfo{pages}{100--103}.
\newblock


\bibitem[Mildenhall et~al\mbox{.}(2021)]%
        {mildenhall2021nerf}
\bibfield{author}{\bibinfo{person}{Ben Mildenhall}, \bibinfo{person}{Pratul~P Srinivasan}, \bibinfo{person}{Matthew Tancik}, \bibinfo{person}{Jonathan~T Barron}, \bibinfo{person}{Ravi Ramamoorthi}, {and} \bibinfo{person}{Ren Ng}.} \bibinfo{year}{2021}\natexlab{}.
\newblock \showarticletitle{Nerf: Representing scenes as neural radiance fields for view synthesis}.
\newblock \bibinfo{journal}{\emph{Commun. ACM}} \bibinfo{volume}{65}, \bibinfo{number}{1} (\bibinfo{year}{2021}), \bibinfo{pages}{99--106}.
\newblock


\bibitem[Mo et~al\mbox{.}(2020)]%
        {mo2020freeze}
\bibfield{author}{\bibinfo{person}{Sangwoo Mo}, \bibinfo{person}{Minsu Cho}, {and} \bibinfo{person}{Jinwoo Shin}.} \bibinfo{year}{2020}\natexlab{}.
\newblock \showarticletitle{Freeze the Discriminator: a Simple Baseline for Fine-Tuning GANs}. In \bibinfo{booktitle}{\emph{CVPR Workshop}}.
\newblock


\bibitem[Mokady et~al\mbox{.}(2022)]%
        {mokady2022self}
\bibfield{author}{\bibinfo{person}{Ron Mokady}, \bibinfo{person}{Michal Yarom}, \bibinfo{person}{Omer Tov}, \bibinfo{person}{Oran Lang}, \bibinfo{person}{Daniel Cohen-Or}, \bibinfo{person}{Tali Dekel}, \bibinfo{person}{Michal Irani}, {and} \bibinfo{person}{Inbar Mosseri}.} \bibinfo{year}{2022}\natexlab{}.
\newblock \showarticletitle{Self-Distilled StyleGAN: Towards Generation from Internet Photos}. In \bibinfo{booktitle}{\emph{ACM SIGGRAPH}}.
\newblock


\bibitem[Nitzan et~al\mbox{.}(2022)]%
        {nitzan2022mystyle}
\bibfield{author}{\bibinfo{person}{Yotam Nitzan}, \bibinfo{person}{Kfir Aberman}, \bibinfo{person}{Qiurui He}, \bibinfo{person}{Orly Liba}, \bibinfo{person}{Michal Yarom}, \bibinfo{person}{Yossi Gandelsman}, \bibinfo{person}{Inbar Mosseri}, \bibinfo{person}{Yael Pritch}, {and} \bibinfo{person}{Daniel Cohen-Or}.} \bibinfo{year}{2022}\natexlab{}.
\newblock \showarticletitle{MyStyle: A Personalized Generative Prior}.
\newblock \bibinfo{journal}{\emph{arXiv preprint arXiv:2203.17272}} (\bibinfo{year}{2022}).
\newblock


\bibitem[Noguchi and Harada(2019)]%
        {noguchi2019image}
\bibfield{author}{\bibinfo{person}{Atsuhiro Noguchi} {and} \bibinfo{person}{Tatsuya Harada}.} \bibinfo{year}{2019}\natexlab{}.
\newblock \showarticletitle{Image generation from small datasets via batch statistics adaptation}. In \bibinfo{booktitle}{\emph{ICCV}}.
\newblock


\bibitem[Ojha et~al\mbox{.}(2021)]%
        {ojha2021few-shot-gan}
\bibfield{author}{\bibinfo{person}{Utkarsh Ojha}, \bibinfo{person}{Yijun Li}, \bibinfo{person}{Cynthia Lu}, \bibinfo{person}{Alexei~A. Efros}, \bibinfo{person}{Yong~Jae Lee}, \bibinfo{person}{Eli Shechtman}, {and} \bibinfo{person}{Richard Zhang}.} \bibinfo{year}{2021}\natexlab{}.
\newblock \showarticletitle{Few-shot Image Generation via Cross-domain Correspondence}. In \bibinfo{booktitle}{\emph{CVPR}}.
\newblock


\bibitem[Oliva and Torralba(2001)]%
        {oliva2001modeling}
\bibfield{author}{\bibinfo{person}{Aude Oliva} {and} \bibinfo{person}{Antonio Torralba}.} \bibinfo{year}{2001}\natexlab{}.
\newblock \showarticletitle{Modeling the shape of the scene: A holistic representation of the spatial envelope}.
\newblock \bibinfo{journal}{\emph{IJCV}} \bibinfo{volume}{42}, \bibinfo{number}{3} (\bibinfo{year}{2001}), \bibinfo{pages}{145--175}.
\newblock


\bibitem[Oord et~al\mbox{.}(2016)]%
        {oord2016conditional}
\bibfield{author}{\bibinfo{person}{Aaron van~den Oord}, \bibinfo{person}{Nal Kalchbrenner}, \bibinfo{person}{Oriol Vinyals}, \bibinfo{person}{Lasse Espeholt}, \bibinfo{person}{Alex Graves}, {and} \bibinfo{person}{Koray Kavukcuoglu}.} \bibinfo{year}{2016}\natexlab{}.
\newblock \showarticletitle{Conditional Image Generation with PixelCNN Decoders}. In \bibinfo{booktitle}{\emph{NeurIPS}}.
\newblock


\bibitem[Oord et~al\mbox{.}(2018)]%
        {oord2018representation}
\bibfield{author}{\bibinfo{person}{Aaron van~den Oord}, \bibinfo{person}{Yazhe Li}, {and} \bibinfo{person}{Oriol Vinyals}.} \bibinfo{year}{2018}\natexlab{}.
\newblock \showarticletitle{Representation learning with contrastive predictive coding}.
\newblock \bibinfo{journal}{\emph{arXiv preprint arXiv:1807.03748}} (\bibinfo{year}{2018}).
\newblock


\bibitem[Parmar et~al\mbox{.}(2022)]%
        {parmar2021cleanfid}
\bibfield{author}{\bibinfo{person}{Gaurav Parmar}, \bibinfo{person}{Richard Zhang}, {and} \bibinfo{person}{Jun-Yan Zhu}.} \bibinfo{year}{2022}\natexlab{}.
\newblock \showarticletitle{On Buggy Resizing Libraries and Surprising Subtleties in FID Calculation}. In \bibinfo{booktitle}{\emph{CVPR}}.
\newblock


\bibitem[Patashnik et~al\mbox{.}(2021)]%
        {patashnik2021styleclip}
\bibfield{author}{\bibinfo{person}{Or Patashnik}, \bibinfo{person}{Zongze Wu}, \bibinfo{person}{Eli Shechtman}, \bibinfo{person}{Daniel Cohen-Or}, {and} \bibinfo{person}{Dani Lischinski}.} \bibinfo{year}{2021}\natexlab{}.
\newblock \showarticletitle{Styleclip: Text-driven manipulation of stylegan imagery}. In \bibinfo{booktitle}{\emph{ICCV}}. \bibinfo{pages}{2085--2094}.
\newblock


\bibitem[Pinkney(2020)]%
        {pinkney2020awesome}
\bibfield{author}{\bibinfo{person}{Justin Pinkney}.} \bibinfo{year}{2020}\natexlab{}.
\newblock \bibinfo{title}{Awesome Pretrained StyleGAN}.
\newblock \bibinfo{howpublished}{https://www.justinpinkney.com/pretrained-stylegan/}.
\newblock


\bibitem[Poole et~al\mbox{.}(2022)]%
        {poole2022dreamfusion}
\bibfield{author}{\bibinfo{person}{Ben Poole}, \bibinfo{person}{Ajay Jain}, \bibinfo{person}{Jonathan~T Barron}, {and} \bibinfo{person}{Ben Mildenhall}.} \bibinfo{year}{2022}\natexlab{}.
\newblock \showarticletitle{Dreamfusion: Text-to-3d using 2d diffusion}.
\newblock \bibinfo{journal}{\emph{arXiv preprint arXiv:2209.14988}} (\bibinfo{year}{2022}).
\newblock


\bibitem[Radenovic et~al\mbox{.}(2018)]%
        {radenovic2018deep}
\bibfield{author}{\bibinfo{person}{Filip Radenovic}, \bibinfo{person}{Giorgos Tolias}, {and} \bibinfo{person}{Ondrej Chum}.} \bibinfo{year}{2018}\natexlab{}.
\newblock \showarticletitle{Deep shape matching}. In \bibinfo{booktitle}{\emph{ECCV}}.
\newblock


\bibitem[Radford et~al\mbox{.}(2021)]%
        {radford2021learning}
\bibfield{author}{\bibinfo{person}{Alec Radford}, \bibinfo{person}{Jong~Wook Kim}, \bibinfo{person}{Chris Hallacy}, \bibinfo{person}{Aditya Ramesh}, \bibinfo{person}{Gabriel Goh}, \bibinfo{person}{Sandhini Agarwal}, \bibinfo{person}{Girish Sastry}, \bibinfo{person}{Amanda Askell}, \bibinfo{person}{Pamela Mishkin}, \bibinfo{person}{Jack Clark}, \bibinfo{person}{Gretchen Krueger}, {and} \bibinfo{person}{Ilya Sutskever}.} \bibinfo{year}{2021}\natexlab{}.
\newblock \showarticletitle{Learning transferable visual models from natural language supervision}. In \bibinfo{booktitle}{\emph{ICML}}.
\newblock


\bibitem[Ramesh et~al\mbox{.}(2022)]%
        {ramesh2022hierarchical}
\bibfield{author}{\bibinfo{person}{Aditya Ramesh}, \bibinfo{person}{Prafulla Dhariwal}, \bibinfo{person}{Alex Nichol}, \bibinfo{person}{Casey Chu}, {and} \bibinfo{person}{Mark Chen}.} \bibinfo{year}{2022}\natexlab{}.
\newblock \showarticletitle{Hierarchical text-conditional image generation with clip latents}.
\newblock \bibinfo{journal}{\emph{arXiv preprint arXiv:2204.06125}} (\bibinfo{year}{2022}).
\newblock


\bibitem[Razavi et~al\mbox{.}(2019)]%
        {razavi2019generating}
\bibfield{author}{\bibinfo{person}{Ali Razavi}, \bibinfo{person}{Aaron van~den Oord}, {and} \bibinfo{person}{Oriol Vinyals}.} \bibinfo{year}{2019}\natexlab{}.
\newblock \showarticletitle{Generating diverse high-fidelity images with vq-vae-2}. In \bibinfo{booktitle}{\emph{NeurIPS}}.
\newblock


\bibitem[Ribeiro et~al\mbox{.}(2020)]%
        {ribeiro2020sketchformer}
\bibfield{author}{\bibinfo{person}{Leo Sampaio~Ferraz Ribeiro}, \bibinfo{person}{Tu Bui}, \bibinfo{person}{John Collomosse}, {and} \bibinfo{person}{Moacir Ponti}.} \bibinfo{year}{2020}\natexlab{}.
\newblock \showarticletitle{Sketchformer: Transformer-based representation for sketched structure}. In \bibinfo{booktitle}{\emph{CVPR}}.
\newblock


\bibitem[Rombach et~al\mbox{.}(2022)]%
        {rombach2021highresolution}
\bibfield{author}{\bibinfo{person}{Robin Rombach}, \bibinfo{person}{Andreas Blattmann}, \bibinfo{person}{Dominik Lorenz}, \bibinfo{person}{Patrick Esser}, {and} \bibinfo{person}{Björn Ommer}.} \bibinfo{year}{2022}\natexlab{}.
\newblock \showarticletitle{High-Resolution Image Synthesis with Latent Diffusion Models}. In \bibinfo{booktitle}{\emph{CVPR}}.
\newblock


\bibitem[Ruiz et~al\mbox{.}(2022)]%
        {ruiz2022dreambooth}
\bibfield{author}{\bibinfo{person}{Nataniel Ruiz}, \bibinfo{person}{Yuanzhen Li}, \bibinfo{person}{Varun Jampani}, \bibinfo{person}{Yael Pritch}, \bibinfo{person}{Michael Rubinstein}, {and} \bibinfo{person}{Kfir Aberman}.} \bibinfo{year}{2022}\natexlab{}.
\newblock \showarticletitle{DreamBooth: Fine Tuning Text-to-image Diffusion Models for Subject-Driven Generation}. In \bibinfo{booktitle}{\emph{arXiv preprint arxiv:2208.12242}}.
\newblock


\bibitem[Saharia et~al\mbox{.}(2022)]%
        {saharia2022photorealistic}
\bibfield{author}{\bibinfo{person}{Chitwan Saharia}, \bibinfo{person}{William Chan}, \bibinfo{person}{Saurabh Saxena}, \bibinfo{person}{Lala Li}, \bibinfo{person}{Jay Whang}, \bibinfo{person}{Emily~L Denton}, \bibinfo{person}{Kamyar Ghasemipour}, \bibinfo{person}{Raphael Gontijo~Lopes}, \bibinfo{person}{Burcu Karagol~Ayan}, \bibinfo{person}{Tim Salimans}, {et~al\mbox{.}}} \bibinfo{year}{2022}\natexlab{}.
\newblock \showarticletitle{Photorealistic text-to-image diffusion models with deep language understanding}.
\newblock \bibinfo{journal}{\emph{Advances in Neural Information Processing Systems}}  \bibinfo{volume}{35} (\bibinfo{year}{2022}), \bibinfo{pages}{36479--36494}.
\newblock


\bibitem[Sangkloy et~al\mbox{.}(2016)]%
        {sangkloy2016sketchy}
\bibfield{author}{\bibinfo{person}{Patsorn Sangkloy}, \bibinfo{person}{Nathan Burnell}, \bibinfo{person}{Cusuh Ham}, {and} \bibinfo{person}{James Hays}.} \bibinfo{year}{2016}\natexlab{}.
\newblock \showarticletitle{The sketchy database: learning to retrieve badly drawn bunnies}.
\newblock \bibinfo{journal}{\emph{ACM TOG}} \bibinfo{volume}{35}, \bibinfo{number}{4} (\bibinfo{year}{2016}), \bibinfo{pages}{1--12}.
\newblock


\bibitem[Sauer et~al\mbox{.}(2021)]%
        {Sauer2021NEURIPS}
\bibfield{author}{\bibinfo{person}{Axel Sauer}, \bibinfo{person}{Kashyap Chitta}, \bibinfo{person}{Jens M{\"{u}}ller}, {and} \bibinfo{person}{Andreas Geiger}.} \bibinfo{year}{2021}\natexlab{}.
\newblock \showarticletitle{Projected GANs Converge Faster}. In \bibinfo{booktitle}{\emph{NeurIPS}}.
\newblock


\bibitem[Sauer et~al\mbox{.}(2022)]%
        {sauer2022stylegan}
\bibfield{author}{\bibinfo{person}{Axel Sauer}, \bibinfo{person}{Katja Schwarz}, {and} \bibinfo{person}{Andreas Geiger}.} \bibinfo{year}{2022}\natexlab{}.
\newblock \showarticletitle{Stylegan-xl: Scaling stylegan to large diverse datasets}. In \bibinfo{booktitle}{\emph{ACM SIGGRAPH}}.
\newblock


\bibitem[Schultz(2020)]%
        {schultz2020freagan}
\bibfield{author}{\bibinfo{person}{Derrick Schultz}.} \bibinfo{year}{2020}\natexlab{}.
\newblock \bibinfo{title}{FreaGAN, undertrained GAN trained on {Frea Buckler's} artwork}.
\newblock \bibinfo{howpublished}{https://twitter.com/dvsch/status/1255885874560225284}.
\newblock


\bibitem[Shen et~al\mbox{.}(2023)]%
        {shen2023hugginggpt}
\bibfield{author}{\bibinfo{person}{Yongliang Shen}, \bibinfo{person}{Kaitao Song}, \bibinfo{person}{Xu Tan}, \bibinfo{person}{Dongsheng Li}, \bibinfo{person}{Weiming Lu}, {and} \bibinfo{person}{Yueting Zhuang}.} \bibinfo{year}{2023}\natexlab{}.
\newblock \showarticletitle{Hugginggpt: Solving ai tasks with chatgpt and its friends in huggingface}.
\newblock \bibinfo{journal}{\emph{arXiv preprint arXiv:2303.17580}} (\bibinfo{year}{2023}).
\newblock


\bibitem[Sivic and Zisserman(2003)]%
        {sivic2003video}
\bibfield{author}{\bibinfo{person}{Josef Sivic} {and} \bibinfo{person}{Andrew Zisserman}.} \bibinfo{year}{2003}\natexlab{}.
\newblock \showarticletitle{Video Google: A text retrieval approach to object matching in videos}. In \bibinfo{booktitle}{\emph{ICCV}}.
\newblock


\bibitem[Smeulders et~al\mbox{.}(2000)]%
        {smeulders2000content}
\bibfield{author}{\bibinfo{person}{Arnold~WM Smeulders}, \bibinfo{person}{Marcel Worring}, \bibinfo{person}{Simone Santini}, \bibinfo{person}{Amarnath Gupta}, {and} \bibinfo{person}{Ramesh Jain}.} \bibinfo{year}{2000}\natexlab{}.
\newblock \showarticletitle{Content-based image retrieval at the end of the early years}.
\newblock \bibinfo{journal}{\emph{IEEE TPAMI}} (\bibinfo{year}{2000}).
\newblock


\bibitem[Socher et~al\mbox{.}(2014)]%
        {socher2014grounded}
\bibfield{author}{\bibinfo{person}{Richard Socher}, \bibinfo{person}{Andrej Karpathy}, \bibinfo{person}{Quoc~V Le}, \bibinfo{person}{Christopher~D Manning}, {and} \bibinfo{person}{Andrew~Y Ng}.} \bibinfo{year}{2014}\natexlab{}.
\newblock \showarticletitle{Grounded compositional semantics for finding and describing images with sentences}.
\newblock \bibinfo{journal}{\emph{Transactions of the Association for Computational Linguistics}}  \bibinfo{volume}{2} (\bibinfo{year}{2014}), \bibinfo{pages}{207--218}.
\newblock


\bibitem[Song et~al\mbox{.}(2021a)]%
        {song2020denoising}
\bibfield{author}{\bibinfo{person}{Jiaming Song}, \bibinfo{person}{Chenlin Meng}, {and} \bibinfo{person}{Stefano Ermon}.} \bibinfo{year}{2021}\natexlab{a}.
\newblock \showarticletitle{Denoising diffusion implicit models}. In \bibinfo{booktitle}{\emph{ICLR}}.
\newblock


\bibitem[Song et~al\mbox{.}(2021b)]%
        {song2020score}
\bibfield{author}{\bibinfo{person}{Yang Song}, \bibinfo{person}{Jascha Sohl-Dickstein}, \bibinfo{person}{Diederik~P Kingma}, \bibinfo{person}{Abhishek Kumar}, \bibinfo{person}{Stefano Ermon}, {and} \bibinfo{person}{Ben Poole}.} \bibinfo{year}{2021}\natexlab{b}.
\newblock \showarticletitle{Score-based generative modeling through stochastic differential equations}. In \bibinfo{booktitle}{\emph{ICLR}}.
\newblock


\bibitem[Tewari et~al\mbox{.}(2020)]%
        {Tewari2020NeuralSTAR}
\bibfield{author}{\bibinfo{person}{A. Tewari}, \bibinfo{person}{O. Fried}, \bibinfo{person}{J. Thies}, \bibinfo{person}{V. Sitzmann}, \bibinfo{person}{S. Lombardi}, \bibinfo{person}{K. Sunkavalli}, \bibinfo{person}{R. Martin-Brualla}, \bibinfo{person}{T. Simon}, \bibinfo{person}{J. Saragih}, \bibinfo{person}{M. Nie{\ss}ner}, \bibinfo{person}{R. Pandey}, \bibinfo{person}{S. Fanello}, \bibinfo{person}{G. Wetzstein}, \bibinfo{person}{J.-Y. Zhu}, \bibinfo{person}{C. Theobalt}, \bibinfo{person}{M. Agrawala}, \bibinfo{person}{E. Shechtman}, \bibinfo{person}{D.~B Goldman}, {and} \bibinfo{person}{M. Zollh{\"o}fer}.} \bibinfo{year}{2020}\natexlab{}.
\newblock \showarticletitle{{State of the Art on Neural Rendering}}.
\newblock \bibinfo{journal}{\emph{Computer Graphics Forum (EG STAR 2020)}} (\bibinfo{year}{2020}).
\newblock


\bibitem[Torralba et~al\mbox{.}(2008)]%
        {torralba2008small}
\bibfield{author}{\bibinfo{person}{Antonio Torralba}, \bibinfo{person}{Rob Fergus}, {and} \bibinfo{person}{Yair Weiss}.} \bibinfo{year}{2008}\natexlab{}.
\newblock \showarticletitle{Small codes and large image databases for recognition}. In \bibinfo{booktitle}{\emph{CVPR}}.
\newblock


\bibitem[Tumanyan et~al\mbox{.}(2023)]%
        {tumanyan2022plug}
\bibfield{author}{\bibinfo{person}{Narek Tumanyan}, \bibinfo{person}{Michal Geyer}, \bibinfo{person}{Shai Bagon}, {and} \bibinfo{person}{Tali Dekel}.} \bibinfo{year}{2023}\natexlab{}.
\newblock \showarticletitle{Plug-and-play diffusion features for text-driven image-to-image translation}. In \bibinfo{booktitle}{\emph{Proceedings of the IEEE/CVF Conference on Computer Vision and Pattern Recognition}}. \bibinfo{pages}{1921--1930}.
\newblock


\bibitem[Unsplash(2022)]%
        {Unsplash}
\bibfield{author}{\bibinfo{person}{Unsplash}.} \bibinfo{year}{2022}\natexlab{}.
\newblock \bibinfo{title}{Unsplash}.
\newblock \bibinfo{howpublished}{\url{https://unsplash.com}}.
\newblock


\bibitem[Wang et~al\mbox{.}(2021)]%
        {wang2021sketch}
\bibfield{author}{\bibinfo{person}{Sheng-Yu Wang}, \bibinfo{person}{David Bau}, {and} \bibinfo{person}{Jun-Yan Zhu}.} \bibinfo{year}{2021}\natexlab{}.
\newblock \showarticletitle{Sketch Your Own GAN}. In \bibinfo{booktitle}{\emph{ICCV}}.
\newblock


\bibitem[Wang et~al\mbox{.}(2022)]%
        {wang2022rewriting}
\bibfield{author}{\bibinfo{person}{Sheng-Yu Wang}, \bibinfo{person}{David Bau}, {and} \bibinfo{person}{Jun-Yan Zhu}.} \bibinfo{year}{2022}\natexlab{}.
\newblock \showarticletitle{Rewriting Geometric Rules of a GAN}.
\newblock \bibinfo{journal}{\emph{ACM TOG}} (\bibinfo{year}{2022}).
\newblock


\bibitem[Wang et~al\mbox{.}(2020)]%
        {wang2020minegan}
\bibfield{author}{\bibinfo{person}{Yaxing Wang}, \bibinfo{person}{Abel Gonzalez-Garcia}, \bibinfo{person}{David Berga}, \bibinfo{person}{Luis Herranz}, \bibinfo{person}{Fahad~Shahbaz Khan}, {and} \bibinfo{person}{Joost van~de Weijer}.} \bibinfo{year}{2020}\natexlab{}.
\newblock \showarticletitle{Minegan: effective knowledge transfer from gans to target domains with few images}. In \bibinfo{booktitle}{\emph{CVPR}}.
\newblock


\bibitem[Wang et~al\mbox{.}(2018)]%
        {wang2018transferring}
\bibfield{author}{\bibinfo{person}{Yaxing Wang}, \bibinfo{person}{Chenshen Wu}, \bibinfo{person}{Luis Herranz}, \bibinfo{person}{Joost van~de Weijer}, \bibinfo{person}{Abel Gonzalez-Garcia}, {and} \bibinfo{person}{Bogdan Raducanu}.} \bibinfo{year}{2018}\natexlab{}.
\newblock \showarticletitle{Transferring gans: generating images from limited data}. In \bibinfo{booktitle}{\emph{ECCV}}.
\newblock


\bibitem[Weiss et~al\mbox{.}(2008)]%
        {weiss2008spectral}
\bibfield{author}{\bibinfo{person}{Yair Weiss}, \bibinfo{person}{Antonio Torralba}, {and} \bibinfo{person}{Rob Fergus}.} \bibinfo{year}{2008}\natexlab{}.
\newblock \showarticletitle{Spectral hashing}. In \bibinfo{booktitle}{\emph{NeurIPS}}.
\newblock


\bibitem[Yu et~al\mbox{.}(2015)]%
        {yu2015lsun}
\bibfield{author}{\bibinfo{person}{Fisher Yu}, \bibinfo{person}{Ari Seff}, \bibinfo{person}{Yinda Zhang}, \bibinfo{person}{Shuran Song}, \bibinfo{person}{Thomas Funkhouser}, {and} \bibinfo{person}{Jianxiong Xiao}.} \bibinfo{year}{2015}\natexlab{}.
\newblock \showarticletitle{Lsun: Construction of a large-scale image dataset using deep learning with humans in the loop}.
\newblock \bibinfo{journal}{\emph{arXiv preprint arXiv:1506.03365}} (\bibinfo{year}{2015}).
\newblock


\bibitem[Yu et~al\mbox{.}(2016)]%
        {yu2016sketch}
\bibfield{author}{\bibinfo{person}{Qian Yu}, \bibinfo{person}{Feng Liu}, \bibinfo{person}{Yi-Zhe Song}, \bibinfo{person}{Tao Xiang}, \bibinfo{person}{Timothy~M Hospedales}, {and} \bibinfo{person}{Chen-Change Loy}.} \bibinfo{year}{2016}\natexlab{}.
\newblock \showarticletitle{Sketch me that shoe}. In \bibinfo{booktitle}{\emph{CVPR}}.
\newblock


\bibitem[Zhang et~al\mbox{.}(2023)]%
        {zhang2023adding}
\bibfield{author}{\bibinfo{person}{Lvmin Zhang}, \bibinfo{person}{Anyi Rao}, {and} \bibinfo{person}{Maneesh Agrawala}.} \bibinfo{year}{2023}\natexlab{}.
\newblock \showarticletitle{Adding conditional control to text-to-image diffusion models}. In \bibinfo{booktitle}{\emph{Proceedings of the IEEE/CVF International Conference on Computer Vision}}. \bibinfo{pages}{3836--3847}.
\newblock


\bibitem[Zhang et~al\mbox{.}(2021)]%
        {zhang2021datasetgan}
\bibfield{author}{\bibinfo{person}{Yuxuan Zhang}, \bibinfo{person}{Huan Ling}, \bibinfo{person}{Jun Gao}, \bibinfo{person}{Kangxue Yin}, \bibinfo{person}{Jean-Francois Lafleche}, \bibinfo{person}{Adela Barriuso}, \bibinfo{person}{Antonio Torralba}, {and} \bibinfo{person}{Sanja Fidler}.} \bibinfo{year}{2021}\natexlab{}.
\newblock \showarticletitle{Datasetgan: Efficient labeled data factory with minimal human effort}. In \bibinfo{booktitle}{\emph{CVPR}}.
\newblock


\bibitem[Zhao et~al\mbox{.}(2020a)]%
        {zhao2020leveraging}
\bibfield{author}{\bibinfo{person}{Miaoyun Zhao}, \bibinfo{person}{Yulai Cong}, {and} \bibinfo{person}{Lawrence Carin}.} \bibinfo{year}{2020}\natexlab{a}.
\newblock \showarticletitle{On leveraging pretrained GANs for generation with limited data}. In \bibinfo{booktitle}{\emph{ICML}}.
\newblock


\bibitem[Zhao et~al\mbox{.}(2020b)]%
        {zhao2020diffaugment}
\bibfield{author}{\bibinfo{person}{Shengyu Zhao}, \bibinfo{person}{Zhijian Liu}, \bibinfo{person}{Ji Lin}, \bibinfo{person}{Jun-Yan Zhu}, {and} \bibinfo{person}{Song Han}.} \bibinfo{year}{2020}\natexlab{b}.
\newblock \showarticletitle{Differentiable Augmentation for Data-Efficient GAN Training}. In \bibinfo{booktitle}{\emph{NeurIPS}}.
\newblock


\bibitem[Zheng et~al\mbox{.}(2017)]%
        {zheng2017sift}
\bibfield{author}{\bibinfo{person}{Liang Zheng}, \bibinfo{person}{Yi Yang}, {and} \bibinfo{person}{Qi Tian}.} \bibinfo{year}{2017}\natexlab{}.
\newblock \showarticletitle{SIFT meets CNN: A decade survey of instance retrieval}.
\newblock \bibinfo{journal}{\emph{IEEE TPAMI}} (\bibinfo{year}{2017}).
\newblock


\bibitem[Zhu et~al\mbox{.}(2021)]%
        {zhu2021barbershop}
\bibfield{author}{\bibinfo{person}{Peihao Zhu}, \bibinfo{person}{Rameen Abdal}, \bibinfo{person}{John Femiani}, {and} \bibinfo{person}{Peter Wonka}.} \bibinfo{year}{2021}\natexlab{}.
\newblock \showarticletitle{Barbershop: Gan-based image compositing using segmentation masks}.
\newblock \bibinfo{journal}{\emph{ACM TOG}} (\bibinfo{year}{2021}).
\newblock


\end{thebibliography}

\clearpage
\begin{figure*} %
    \centering %
    \includegraphics[width=0.85\linewidth]{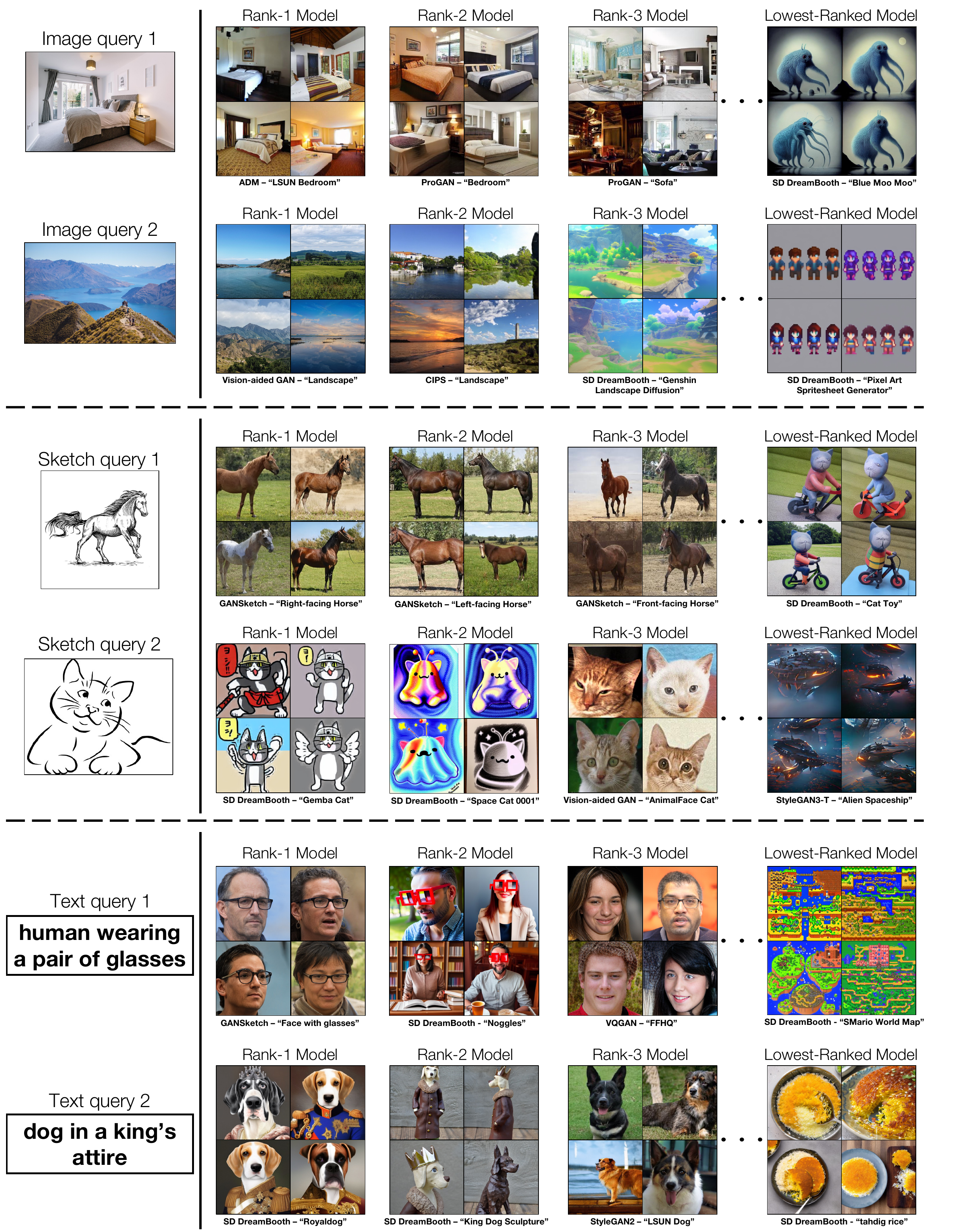}
    \caption{\textbf{Qualitative results of model retrieval.}  \textit{Top row (image query):}  Our algorithm retrieves matching models of indoor scenes and landscapes. \textit{Middle row (sketch query):} Animal sketches retrieve other animals, and more simplistic sketches (e.g. the cat sketch) tend to retrieve cartoonish models. \textit{Bottom row (text query):} We retrieve relevant models matching the textual description. \camready{Photos by \href{https://pixabay.com/users/stubaileyphoto-19245286/}{Stuart Bailey} and \href{https://pixabay.com/users/timbri97-25670575/}{@timbri97} on \href{https://pixabay.com/}{Pixabay}. Sketches by \href{https://pixabay.com/users/dharmah-18954604/}{Damaris Dharmah} and \href{https://pixabay.com/users/openclipart-vectors-30363/}{@OpenClipart-Vectors} on \href{https://pixabay.com/}{Pixabay}.}
    }%
    \lblfig{results_main}%
\end{figure*}
\clearpage

\clearpage
\begin{figure*} %
    \centering %
    \includegraphics[width=0.84\linewidth]{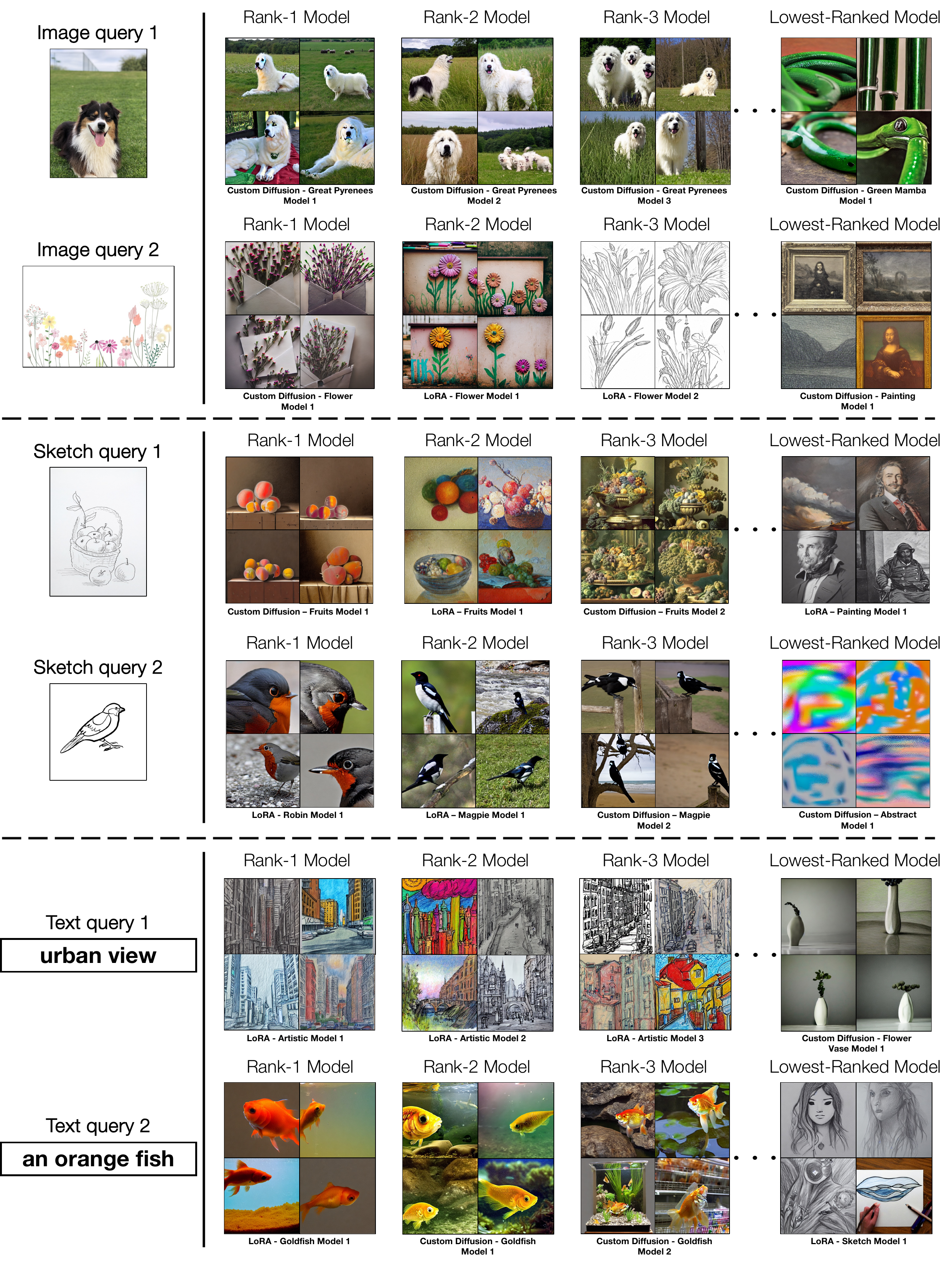}
    \caption{\textbf{Qualitative results of model retrieval on Synthetic Model Zoo.}  \textit{Top row (image query):}  Our algorithm retrieves models of the dog breed Great Pyrenees and artistic flowers, given matching image queries. \camready{\textit{Middle row (sketch query):} Given sketches of a fruit basket and a bird, our algorithm retrieves models that generate a collection of fruits, and other birds, respectively.} \textit{Bottom row (text query):} We can also retrieve relevant models with a textual description. \camready{Images by \href{https://pixabay.com/users/drcaringola-9226780/}{@drcaringola} and \href{https://pixabay.com/users/owantana-3064916/}{@Owantana} on \href{https://pixabay.com/}{Pixabay}. Sketches by \href{https://pixabay.com/users/n-region-6314823/}{@N-region} and \href{https://pixabay.com/users/tilixia-26202716/}{@Tilixia} on \href{https://pixabay.com/}{Pixabay}.}
    }%
    \lblfig{result_additional_synth}%
\end{figure*}
\clearpage

\section*{Appendix}
\appendix

\myparagraph{Overview.}
In \refsec{analysis}, we run additional tests that reveal our method's retrieval score distribution, variance, run-time, and memory profile. %
In \refsec{details}, we describe the Generative Model Zoo details. In \refsec{derive}, we derive our \texttt{\mc} and \texttt{\firstmom} model retrieval baselines from a probabilistic perspective. \refsec{society} discusses potential negative social impacts and social benefits associated with the applications of our model retrieval algorithm. Our code, data, and models will be available upon publication. 

\section{Additional Analysis}\lblsec{analysis}

\myparagraph{Cross Validation on Internet Model Zoo.} We perform 5-fold cross-validation on the Internet Model Zoo and find that fine-tuning improves model retrieval accuracy compared to pre-trained counterparts. We show results in \reftbl{cross_validation_internet_zoo}. The finding is consistent with when we run cross-validation on the Synthetic Model Zoo (\reftbl{image_retrieval_results}).

\begin{figure*}
    \centering
    \includegraphics[width=1.0\linewidth]{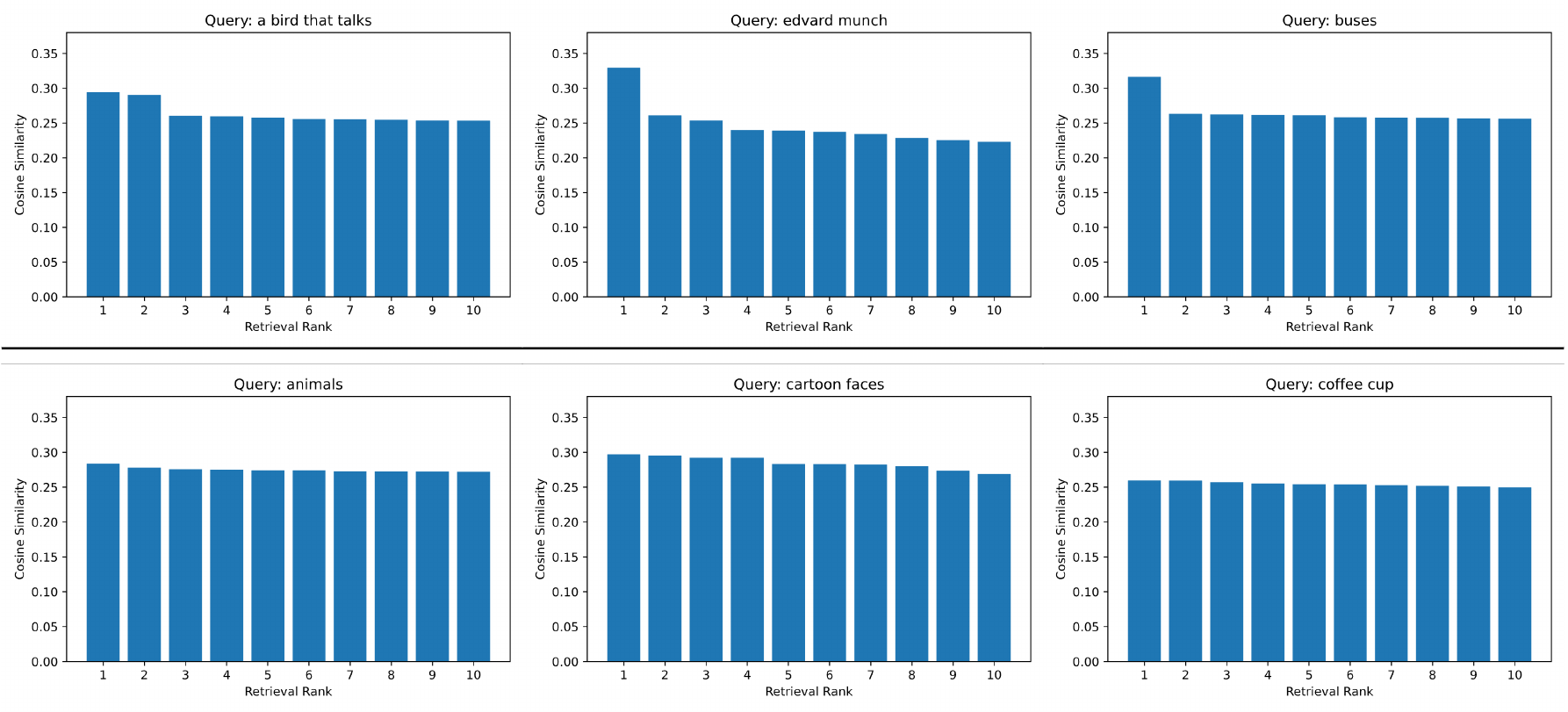}
    \caption{{\bf Similarity score drop-off.} \textit{Top:} \texttt{\firstmom} similarity scores of more specific queries like ``a bird that talks'', ``edvard munch'', or ``buses'' show a clear drop between relevant and irrelevant models. In the first case, the parrot model was ranked top, closely followed by the bird model. In ``edvard munch'', the StyleGAN-NADA Edvard Munch-style face model was ranked top. Subsequent ranks consist of other photorealistic and artistic face models. In ``buses'', the ProGAN bus model was ranked top. Subsequent ranks consisted of models that generate trains, faces, and bridges. \textit{Bottom:} for more general queries like ``animals'' or ``animated faces'', or a query like ``coffee cup'' that does not have any good matches, the similarity scores appear more uniform. In ``animals'', the 10 retrieved models generate animals, including sheep, dogs, cows, horses, cats, and giraffes. In ``cartoon faces'', we find 8 painting/sketch/cartoon-style face models and 2 photorealistic face models. In ``coffee cup'', since our collection does not include any highly relevant models, we find a broad variety of models. The retrieved models consist of a ``bottle'' model, a modern art painting model, a cat model, a ``trypophobia'' model, two modern art painting models, two anime/cartoon character models, and two face models.}
    \label{fig:ret_score}
\end{figure*}

\myparagraph{Distribution of retrieval scores.}
We plotted the top-1-to-10 retrieval scores with the \texttt{\firstmom} method (eq. \ref{eq:first_second}) for a few text queries in \reffig{ret_score}. For more specific queries like ``a bird that talks'' (describing the parrot, a type of bird), ``edvard munch'' (a painter's name), or ``buses'', the \texttt{\firstmom} similarity scores show a clear drop between relevant and irrelevant models. On the other hand, more general queries like ``animals'' or ``animated faces'' can have many relevant models, so the similarity scores appear more uniform. If a query has no good matches, as is the case with ``coffee cup'', the similarity scores also appear uniform.

\myparagraph{Evaluation with mean average precision.}
\reftbl{ablation_mAP} reports the performance of baseline and our fine-tuned methods for model retrieval using mean average precision as the evaluation metric.

\myparagraph{Run-time and memory analysis.}
We profile our methods' running time and memory usage on 259 generative models as well as a much greater number of models in simulation. To simulate models, we sample the \texttt{\firstmom} model statistics and \texttt{\mc} samples from a 512-dimensional normal distribution that correspond to points in the CLIP feature space, and the second moment statistics are generated as a unit covariance matrix with a small, uniform, symmetric noise added. We store the statistics needed for the respective method in the GPU's VRAM during model retrieval. We run the following tests on a machine equipped with an AMD Threadripper 3960X and NVIDIA RTX A5000 running Pytorch 1.11.0, and report the runtimes in \reftbl{time_complexity}.

During model retrieval, extracting query features takes constant time regardless of the number of models to retrieve. In the case of text, extracting CLIP features takes 5.02 ms and 0.36 GB of VRAM. Even for one million models, the \texttt{\firstmom} based scoring method only takes 0.19 ms and 1.93 GB of VRAM. When handling image-to-model or sketch-to-model retrieval, extracting query features with CLIP takes 6.75 ms and 0.36 GB of VRAM. Compared to \texttt{\firstmom}, the \texttt{\gauden} method requires extra memory to store the models' covariance matrices. With $10,000$ models, \texttt{\gauden} takes 0.49ms and 9.85 GB of VRAM. Switching to the \texttt{\firstmom} method offers a 3.06 $\times$ speed up and a 458 $\times$ reduction in memory consumption while achieving a similar accuracy (see \reftbl{image_retrieval_results}). Using the fine-tuned \texttt{\firstmom} improves accuracy, but roughly doubles the running time due to the matrix multiplication involved. Using the fine-tuned \texttt{\gauden}  method also improves accuracy, but comes at no additional computational costs since the covariance matrices can be multiplied with the transformation and cached into memory. The transformation $A_\psi$ is a $512 \times 512$ matrix and takes negligible memory to store.

\begin{table}[!t]
\caption{\textbf{Cross validation on Internet Model Zoo.} We fine-tune different combinations of scoring function and transformation matrix parameterization with 5-fold cross-validation on the Internet Model Zoo. The best-performing parameterization of $A_\psi$ for each scoring method by modality is marked with a ``$\star$''.}
\resizebox{\linewidth}{!}{
\begin{tabular}{llrrrr}
\toprule
      &     &  & & \multicolumn{2}{c}{Mean Testing Accuracy}  \\
      \cmidrule(lr){5-6}
      &     & Modality & $A\psi$  & Top-1 (FT)     & Top-1 (Pre-trained) \\ 
      \midrule \midrule

\multirow{12}{*}{\rotatebox[origin=c]{90}{ \begin{tabular}{@{}c@{}} \textbf{Internet Zoo}\end{tabular} }}
& CLIP+\firstmom        & Image & Diagonal & 0.92 &	0.91  \\
& CLIP+\firstmom        & Image & $\star$Triangular & \textbf{0.94}  &	0.91  \\
& CLIP+\firstmom        & Image & Full &  0.92 &	0.91  \\
& CLIP+\gauden        & Image & $\star$Diagonal & \textbf{0.95}	&	0.94 \\
& CLIP+\gauden        & Image & Triangular & 0.93	&	0.94 \\
& CLIP+\gauden        & Image & Full & 0.92 	&	0.94 \\
\cmidrule{2-6}
& CLIP+\firstmom        & Sketch & Diagonal &	0.60 &	0.55  \\
& CLIP+\firstmom        & Sketch & $\star$Triangular & \textbf{0.75} &	0.55  \\
& CLIP+\firstmom        & Sketch & Full & 0.72  &	0.55  \\
& CLIP+\gauden        & Sketch & $\star$Diagonal &	\textbf{0.57} &	0.35 \\
& CLIP+\gauden        & Sketch & Triangular & 0.46	&	0.35 \\
& CLIP+\gauden        & Sketch & Full &	0.40  &	0.35 \\
\bottomrule
\end{tabular}}
\lbltbl{cross_validation_internet_zoo}

\end{table}

\begin{table}[!t]
\caption{\textbf{mAP numbers.} We report the performance of baseline and our fine-tuned methods for model retrieval using mean average precision as the evaluation metric. For top-K accuracy numbers, see main text \reftbl{image_retrieval_results}.}
\resizebox{0.7\linewidth}{!}{
\begin{tabular}{lcc}
\toprule
&   \textbf{Image (Gen.)} & \textbf{Sketch}    \\
\midrule \midrule
CLIP+Monte-Carlo (2.4K) & 0.88   & 0.45\\
CLIP+best-k (2.4K, k=1)  & 0.87  & 0.41\\
CLIP+best-k (2.4K, k=10)  & 0.88 & 0.45\\
\cmidrule(l{6pt}r{6pt}){1-3}
CLIP+\firstmom  &0.85   &0.45 \\
CLIP+\firstmom~(FT)  &0.87  & \textbf{0.61}  \\
CLIP+\gauden  & 0.89   & 0.29  \\
CLIP+\gauden~(FT) & \textbf{0.90}   & 0.42 \\
\bottomrule
\end{tabular}}
\lbltbl{ablation_mAP}

\end{table}

\begin{table}[!t]
\caption{\textbf{Model retrieval running time.} The table shows the running time of different retrieval methods. OOM stands for ``out of memory'' on a 24GB-VRAM GPU and that the retrieval cannot be done in a single pass. After we get the score for each model, we additionally run \texttt{torch.argsort} to get the best matches, which takes 0.03, 0.15, and 0.12 ms for 259, 10K, and one million models, respectively (not shown in the table).}
\resizebox{\linewidth}{!}{
\begin{tabular}{llcrrr}
\toprule
     &  &   \multirow{2}{*}{\shortstack[c]{Feature\\Extraction}}  & \multicolumn{3}{c}{Model Scoring} \\ \cmidrule(lr){4-6}
     &  & & 259 & 10K & 1M\\ \midrule  \midrule
\multirow{3}{*}{\textbf{Text}} & Monte-Carlo (2.4K) & \multirow{3}{*}{5.02ms} & 0.19ms & OOM & OOM\\

& \firstmom & & \textbf{0.09ms} & \textbf{0.16ms} & \textbf{0.21ms}\\
& \firstmom (FT) & & 0.17ms & {0.30ms} & {0.33ms}\\
\midrule
\multirow{5}{*}{\textbf{Image}} &  Monte-Carlo (2.4K) & \multirow{5}{*}{6.75ms} & 0.19ms & OOM & OOM\\
  &  \gauden & & 0.45ms & 0.49ms & OOM \\
  &  \gauden (FT) & & 0.45ms & 0.49ms & OOM \\
  &   \firstmom &  & \textbf{0.06ms} & \textbf{0.15ms} & \textbf{0.19ms} \\
  &   \firstmom (FT) &  &0.15ms & 0.28ms & 0.32ms \\

\bottomrule
\end{tabular}}
\lbltbl{time_complexity}
\end{table}

After getting the retrieval scores, the time required for sorting and selecting the most relevant models is comparatively short on a GPU. For one million models, sorting takes 0.12ms and 0.062 GB of VRAM. When we use the \texttt{\firstmom} method, we enable a user to retrieve models from a 1-million-model collection with a text, sketch, or image query in under 10 ms.

\begin{figure*}
    \centering
    \includegraphics[width=1.0\textwidth]{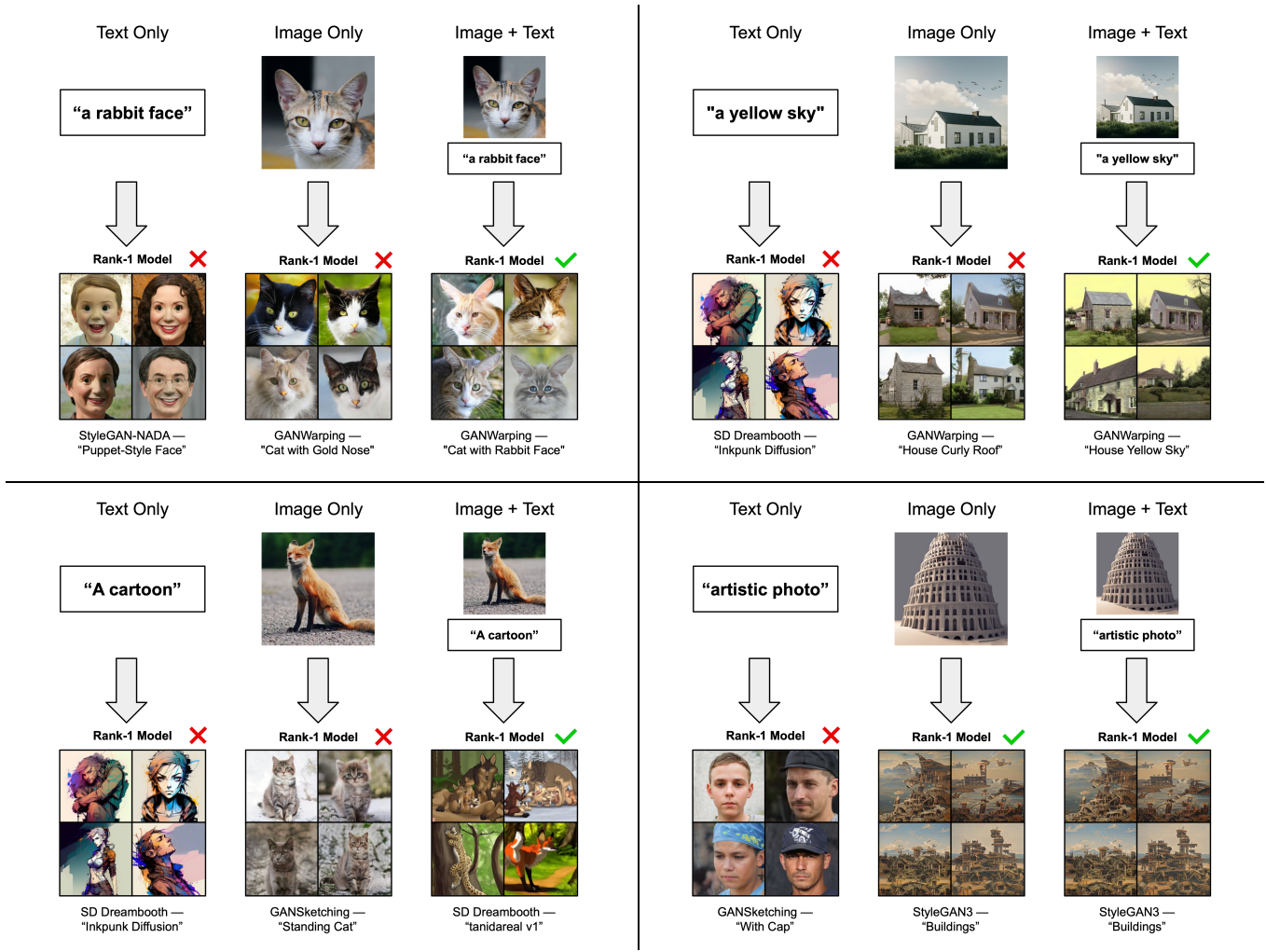}
    \caption{{\bf Evaluation over multi-modal queries.} We show the qualitative results of retrieval using text only, image only, and image + text as the query. We search over a subset of the Internet Model Zoo, specifically, we selected 36 models whose descriptions can be split into text and image. We can utilize direction from both text and image to retrieve models more consistently. Photos by \href{https://unsplash.com/@aalochak}{Himanshu Choudhary}, \href{https://unsplash.com/@alex_andrews}{Alexander Andrews}, and \href{https://unsplash.com/@obkim}{Qijin Xu} on \href{https://unsplash.com/}{Unsplash}. Rendering by \href{https://blenderartists.org/u/ForgottenWorld}{@ForgottenWorld} on \href{https://blenderartists.org/}{Blender Artists} (bottom right).}
    \label{fig:multimodal_appendix}
\end{figure*}

\begin{table}[!t]
\caption{\textbf{Model search with fewer-sample statistics.} We run model search while using fewer samples to compute the feature-space image statistics for each model at test-time. ``ALL" refers to using all 50K or 2,400 available samples depending on the model, which is the default (see \refsec{details}). In in other tests we randomly sample a subset of given size.}
\resizebox{\linewidth}{!}{
\begin{tabular}{llrrrr}
\toprule
      &     &  & & \multicolumn{2}{c}{Mean Testing Accuracy}  \\
      \cmidrule(lr){5-6}
      &     & Modality & Stats Sample Size  & Top-1 (FT)     & Top-1 (Pre-trained) \\ 
      \midrule \midrule

\multirow{12}{*}{\rotatebox[origin=c]{90}{ \begin{tabular}{@{}c@{}} \textbf{Internet Zoo}\end{tabular} }}
& CLIP+\gauden       & Image & \textit{ALL} & 0.85	& 0.84 \\
& CLIP+\gauden       & Image & 2,400 & 0.85	& 0.84 \\
& CLIP+\gauden       & Image & 100 & 0.79	&	0.78 \\
& CLIP+\gauden       & Image & 10 &  0.68	&	0.66 \\
\cmidrule{2-6}
& CLIP+\firstmom       & Sketch & \textit{ALL} &	0.49 & 0.32 \\
& CLIP+\firstmom       & Sketch & 2,400 &	0.49 & 0.32 \\
& CLIP+\firstmom       & Sketch & 100 & 0.46 &	0.30  \\
& CLIP+\firstmom        & Sketch & 10 &  0.40 &	0.27  \\
\bottomrule
\end{tabular}}
\lbltbl{ablation_fewer_samples}

\end{table}

\myparagraph{Model Search with Fewer-sample Model Statistics}
To study how well model search works if model statistics were gathered from fewer generated images, at test-time, we use only a portion of the generated images to compute the feature-space mean and covariance statistics, while keeping all else the same. We show the results in \reftbl{ablation_fewer_samples}. Overall, while a greater sample size helps, we get adequate performance with as few as 100 samples.

\myparagraph{Evaluation over multi-modal queries}
To test retrieval with multi-modal (image + text) queries, we select a subset of 36 models from the Internet Model Zoo whose description can be split into some text and a generic image (e.g., "purple fur" + image of a regular horse, for a purple horse model). Our CLIP+\texttt{\firstmom} method results in 0.47 and 0.86 for top-1 and top-5 accuracies. The qualitative results are presented in \reffig{multimodal_appendix}.

\begin{figure}
    \centering
    \includegraphics[width=1.0\linewidth]{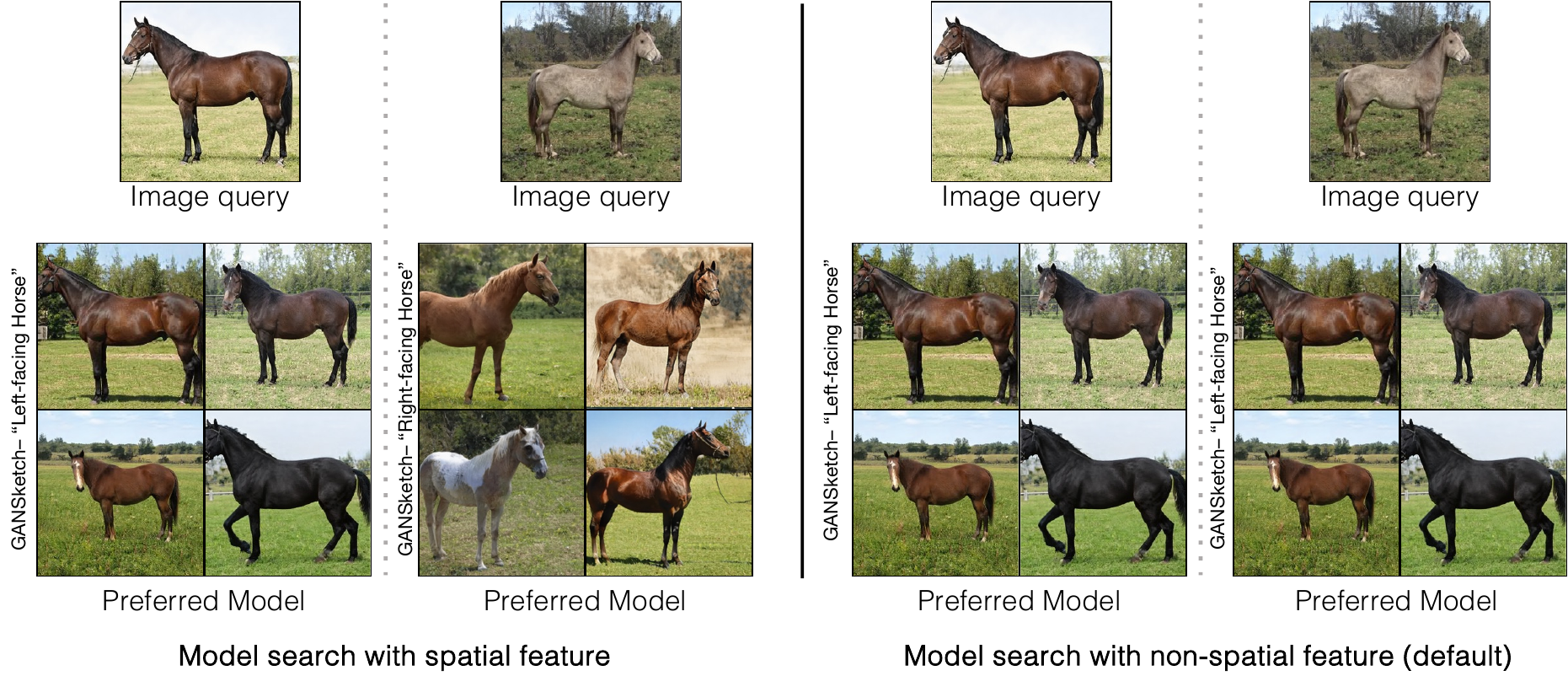}
    \caption{{\bf Refined model retrieval with respect to the spatial layout of the query.} We compare image-based model retrieval using the last layer 7x7 spatial features from ConvNext~\cite{liu2022convnet} pretrained on ImageNet-22k ({\firstmom} method) vs. using global CLIP features ({\gauden (FT)} method). In the last column, the search with global features fails to retrieve the model facing the same direction as the query, whereas the search with spatial features retrieves the correct model in both cases shown. }
    \label{fig:spatial}
\end{figure}

\myparagraph{Refined search using spatial features}
As shown in Figure 6 in the main paper, our method sometimes fails when there is a directional dependency, e.g., left-facing vs. right-facing horse models. To this end, we experiment with spatial features. Given 50 image queries of left- and right-facing horses (identical images, flipped), we take 7x7 spatial features from the last layer of ConvNext~\cite{liu2022convnet} pre-trained on ImageNet-22k and see if each query retrieves the left- or the right-facing-horse model. The horse model facing the same direction as the query is favored 100$\%$ of the time using spatial features and the \texttt{\firstmom} method. In contrast, \texttt{\firstmom} (pre-trained), \texttt{\firstmom} (fine-tuned), \texttt{\gauden} (pre-trained), and \texttt{\gauden} (fine-tuned) methods favor the correct model $52\%, 52\%, 49\%$ and $54\%$ of the time, respectively. \reffig{spatial} shows a qualitative comparison between using spatial features vs. global CLIP features when retrieving left and right-facing horse models with our \texttt{\gauden} (fine-tuned) method. Model search with global CLIP features retrieves the right-facing horse model even though the query depicted a left-facing horse (\reffig{spatial}, last column), where we retrieve the correct model with spatial features.

\section{Dataset Details}\lblsec{details}

\myparagraph{Distribution of Models within Generative Model Zoo.} The 259 models in Internet Model Zoo consist of 1 ADM model~\cite{dhariwal2021diffusion}, 3 CIPS~\cite{anokhin2020image}, 2 DDIM~\cite{song2020denoising}, 2 DDPM~\cite{ho2020denoising}, 1 FastGAN~\cite{liu2020towards}, 18 GANWarping \cite{wang2022rewriting}, 14 GANSketching~\cite{wang2021sketch}, 31 ProGAN~\cite{karras2018progressive}, 7 Self-Distilled GAN~\cite{mokady2022self}, 2 StyleGAN-XL~\cite{sauer2022stylegan}, 2 VQGAN~\cite{esser2021taming}, 10 Vision-Aided GAN~\cite{kumari2021ensembling}, 21 StyleGAN-2~\cite{karras2020analyzing}, 17 StyleGAN-3~\cite{Karras2021alias}, 20 StyleGAN-NADA~\cite{gal2021stylegan} models, and 108 DreamBooth~\cite{ruiz2022dreambooth} fine-tuned Stable-Diffusion models. 

\myparagraph{Model statistics calculation.}
For the Internet Model Zoo, we use 50K generated images to pre-calculate the sample mean and covariance, except for the 108 DreamBooth fine-tuned Stable-Diffusion models where we use 2,400 generated images. In the case of Synthetic Model Zoo, we sample 2400 images to calculate the model statistics, as it consists of 1000 customized diffusion models sampling from which is computationally inefficient. We always add a small factor ($1e^{-4})$ of the identity matrix to the covariance estimate to improve numerical stability. For ImageNet fine-tuned models, we generate class-specific prompts used to sample images using ChatGPT with the instruction: ``Generate 32 captions for images containing a <class-name>. The caption should also contain the word <class-name>''. For artistic fine-tuned models, we use the following two prompts: ``a painting in the style of <specific art style>'' and ``a drawing in the style of <specific art style>''.

\myparagraph{Generating Image and Sketch Queries.}
For every model, we generate 50 more images with the same hyperparameters as used to generate images that are used to compute its distribution statistics. We then create sketches queries by synthesizing sketches from the image queries using the method of Chan \emph{et al.}~\shortcite{chan2022learning}.

\section{Derivation for Text-Based Model Retrieval}\lblsec{derive}
\myparagraph{Monte Carlo Estimation.}
Given a text query $\qry$ and a generative model $p(\im | \g)$ capturing a distribution of images $\im$, we want to estimate the conditional probability $p(\qry | \g)$.

\begin{equation}
\begin{aligned}
    p(\qry | \g) &= \int p(\qry, \im | \g) d\im = \int p(\qry | \im) p(\im | \g) d\im \\
                 &= \int \frac{p(\im | \qry)p(\qry)}{p(\im)} p(\im | \g) d\im \propto \int \frac{p(\im | \qry)}{p(\im)} p(\im | \g) d\im
\end{aligned}
\label{eq:sup_general_eq}
\end{equation}

Here we assume conditional independence between query $\qry$ and model $\g$ given image $x$ (i.e., $p(\qry | \im, \g) = p(\qry | \im)$), since the same image $\im$ should match the same set of queries regardless of the model choice. We apply Bayes' rule in the second line of \refeq{sup_general_eq} to get the final expression. In \refeq{sup_general_eq}, a text query $\qry$ may correspond to multiple possible image matches $p(\im|\qry)$, and we estimate the term $\frac{p(\im|\qry)}{p(\im)}$ using cross-modal similarities. In fact, this expression is proportional to the score function $f$ in contrastive learning (e.g., InfoNCE~\cite{oord2018representation}), where $f(\im, \qry) \propto \frac{p(\im|\qry)}{p(\im)}$. Since CLIP~\cite{radford2021learning} is trained on a text-image retrieval task with the InfoNCE loss, we can directly apply the pre-trained CLIP model to simplify \refeq{sup_general_eq}.

\begin{equation}
    \begin{aligned}
    \max_{\gset} \int \imdist f(\im, \qry) d\im = \mathbb{E}_{\im \sim \imdist}\left[f(\im, \qry)\right].
    \end{aligned}
    \label{eq:clip_score_sup}
\end{equation}
We recall that CLIP consists of an image encoder $\clipimex$ and a text encoder $\cliptxtex$, and it is trained with a score function based on cosine similarity: 
\begin{equation}
\label{eq:text_exp_unnorm_sup}
    f(\im, \qry) = \exp\left(\frac{{\imclip}^T\txtclip_\qry}{\tau(||{\imclip}|| \cdot ||\txtclip_\qry||)}\right) = \exp\left(\frac{\nmimclip^T\nmtxtclip}{\tau}\right),
\end{equation}
\noindent where $\imclip \coloneqq \clipimex(x)$ and $\txtclip_\qry \coloneqq \cliptxtex(\qry)$ are the image and text feature from CLIP, respectively. $\nmimclip = \frac{\imclip}{||\imclip||}$ and $\nmtxtclip = \frac{\txtclip_\qry}{||\txtclip_\qry||}$ are the normalized features. Hence, \refeq{text_exp_unnorm_sup} can be written precisely as:
\begin{equation}
    \begin{aligned}
    \max_{\gset} \mathbb{E}_{\imclip \sim p(\imclip | \g)}\left[\exp\left(\frac{\nmimclip^T\nmtxtclip}{\tau}\right)\right]
    \end{aligned}
    \label{eq:text_exp_sup}
\end{equation}

Even though we cannot compute the expectation inside \refeq{text_exp_sup} in closed form, we can estimate this quantity by sampling from $p(\imclip | \g)$ and computing an empirical average. We call this method \texttt{\mc}. To further speed up model retrieval, we provide an approximation for \refeq{text_exp_sup} -- the \texttt{\firstmom} method. 

\myparagraph{\firstmom\ Method.}
We derive a point estimation at the first moment of the model distribution. Specifically, we estimate the mean of the normalized CLIP image features. (Since CLIP features are learned to be magnitude-invariant, we normalize CLIP features in all of our methods.) Here, we can approximate the feature distribution using a Dirac delta function, where $p(\imclip | \g) \sim \delta(\imclip - \mu_\mi)$. $\mu_\mi$ is the first moment of the distribution, where $\mu_\mi =  \mathbb{E}_{\imclip \sim p(\imclip | \gi)}\left[h\right]$. With this approximation, we can rewrite Equation~\ref{eq:text_exp_sup} as:

\begin{equation}
    \begin{aligned}
    &\max_{\gset} \exp\left(\frac{\Tilde{\mu_\mi}^T\nmtxtclip}{\tau}\right)\\
    \text{where}\;\; &\mu_\mi =  \left(\mathbb{E}_{\imclip \sim p(\imclip | \gi)}\left[\nmimclip\right]\right); \;  \Tilde{\mu_\mi} = \frac{\mu_\mi}{||\mu_\mi||}
    \end{aligned}
    \label{eq:first_one}
\end{equation}

Since the exponential mapping and temperature mapping are monotonically increasing functions, we can further simplify the expression to be our \firstmom\ method: 

\begin{equation}
    \begin{aligned}
    &\max_{\gset} \Tilde{\mu_\mi}^T\nmtxtclip\\
    \text{where}\;\; \mu_\mi =  &\left(\mathbb{E}_{\imclip \sim p(\imclip | \gi)}\left[\nmimclip\right]\right); \;  \Tilde{\mu_\mi} = \frac{\mu_\mi}{||\mu_\mi||}
    \end{aligned}
    \label{eq:first_second}
\end{equation}

\section{Societal Impact}\lblsec{society}

Our method is a step towards making the plethora of generative models cropping up nowadays available to users with ease. However, model search could be used in negative ways: simplifying access to many models may make it easier for malicious actors to create fake content that is difficult to distinguish from real photos. The efficient search may also make it easier to distribute or locate private, offensive, or other damaging information from within a large set of pre-trained models. For that reason, we believe it is important for a model search index to maintain a code of conduct for users and contributors. On the other hand, developing effective model search methods has potential benefits: it may help reduce errors that result from using the wrong model for a problem; it may be a resource for identifying sources of fakes; and it potentially reduces the resources consumed by training redundant models when a good model for a problem already exists.

\end{document}